%% file: main.tex
\documentclass[journal]{IEEEtran}
\usepackage{amsmath,amsfonts}
\usepackage{algorithmic}
\usepackage{algorithm}
\usepackage{array}
\usepackage[caption=false,font=footnotesize,labelfont=rm,textfont=rm]{subfig}
\usepackage{textcomp}
\usepackage{stfloats}
\usepackage{url}
\usepackage{verbatim}
\usepackage{graphicx}
\usepackage{cite}
\usepackage{graphicx}
\usepackage{pict2e}
\usepackage{multirow}
\usepackage{bm}
\usepackage{booktabs}
\usepackage{threeparttable}
\usepackage{amssymb}
\usepackage{amsthm}

\theoremstyle{plain}

\newtheorem{proposition}{Proposition}
\newtheorem{lemma}{Lemma}

\theoremstyle{definition}

\newtheorem{assumption}{Assumption}
\theoremstyle{remark}

\usepackage[colorlinks,linkcolor=black,anchorcolor=black,citecolor=black,urlcolor=black]{hyperref}
\usepackage[utf8]{inputenc}
\usepackage[T1]{fontenc}

\begin{document}

\title{Adaptive Momentum Benefits Score-based Generative Models}

\author{Ziqing Wen, Ping Luo, Jiahuan Wang, Xiaoge Deng, Tao Sun$^*$, and Dongsheng Li
\thanks{Manuscript received. \textit{Corresponding author: Tao Sun.}}
\thanks{Ziqing Wen, Ping Luo, Jiahuan Wang, Tao Sun, Dongsheng Li are with the National Key Laboratory of Parallel and Distributed Computing, National University of Defense Technology, Changsha, 410073, China (e-mail: 
\href{zqwen@nudt.edu.cn}{zqwen@nudt.edu.cn}; 
\href{luoping@nudt.edu.cn}{luoping@nudt.edu.cn}; 
\href{wangjiahuan@nudt.edu.cn}{wangjiahuan@nudt.edu.cn};
\href{suntao.saltfish@outlook.com}{suntao.saltfish@outlook.com}; \href{dsli@nudt.edu.cn}{dsli@nudt.edu.cn}).}
\thanks{Xiaoge Deng is with the Intelligent Game and Decision Lab, Beijing, China (e-mail: \href{dengxg@ustc.edu}{dengxg@ustc.edu}).}
\thanks{Tao Sun is supported in part by the National Natural Science Foundation of China
(Grant Nos.  62522610,62376278), and NUDT Foundational Research Funding (JS25-02).}
}

\markboth{Journal of \LaTeX\ Class Files,~Vol.~14, No.~8, August~2021}%
{Shell \MakeLowercase{\textit{et al.}}: A Sample Article Using IEEEtran.cls for IEEE Journals}

\IEEEpubidadjcol

\maketitle

\begin{abstract}
Score-based Generative Models (SGMs) have achieved remarkable success in generation tasks by establishing a diffusion process that gradually perturbs real data into Gaussian noise and then learning a reverse process to reconstruct the data from noise. However, conventional denoising methods, such as Langevin dynamics and numerical stochastic differential equation solvers, rely on randomness but suffer from slow generation speeds due to the high number of score function evaluations. To address this, we propose a plug-and-play Markov adaptive momentum sampling method, inspired by the strong connection between the SGMs' sampling process and stochastic gradient descent optimization. Our approach accelerates the transformation process without introducing additional hyperparameters. Theoretically, we establish a rigorous foundation by analyzing the relationship between the distribution of samples generated by our method and the original data distribution. This provides a novel perspective on the analysis of sampling distributions. Empirically, our method demonstrates superior efficiency, generating high-quality images/graphs in significantly fewer sampling steps—achieving a speedup of 2× to 5×—while maintaining or surpassing baseline performance on various image and graph generation benchmarks. These results underscore the effectiveness of adaptive momentum in enhancing the efficiency of SGMs.
\end{abstract}

\begin{IEEEkeywords}
Score-based generative models, adaptive sampling, optimization, image/graph generation.
\end{IEEEkeywords}
\section{Introduction}
    \IEEEPARstart{D}{iffusion} models~\cite{ho2020denoising,song2021scorebasedsde}, owing to their strong capacity for learning complex data distributions, have achieved remarkable progress and found applications across diverse domains~\cite{wang2025Diffusion_tip,Bordin2025diffusion_colorguide_tip}.  

    Two prominent classes of diffusion models share the principle of progressively corrupting training data with noise and then learning to reverse this corruption in order to synthesize new samples. The first approach, Score Matching with Langevin Dynamics (SMLD)~\cite{Hyvärinen2009,vincent2011connection}, trains a network to estimate the score function (i.e., the gradient of the log-density of the data distribution). Using these estimated scores together with Langevin Dynamics (LD)~\cite{roberts1996exponential,welling2011bayesian}, new samples can be generated by iteratively following the gradient field.  
    
    A second influential approach is Denoising Diffusion Probabilistic Models (DDPM)~\cite{ho2020denoising}, which formulates the reverse process probabilistically, enabling high-quality image generation. Building on these foundations, Song et al.~\cite{song2021scorebasedsde} proposed a unified score-based stochastic differential equation (SDE) framework that generalizes both diffusion and score-based models. Throughout this paper, we collectively refer to these methods as Score-based Generative Models (SGMs).  
    
    SGMs have demonstrated state-of-the-art performance in a wide range of generative tasks, including image synthesis~\cite{ho2020denoising,song2021scorebasedsde}, video generation~\cite{Ho_vedio_diffusion}, and graph generation~\cite{Jo2022ScorebasedGMgraph,Niu2020PermutationIG,vignac2023digress}. Nevertheless, their success comes at a high computational cost: generating high-fidelity images often requires more than 1,000 sampling steps, leading to significantly slower inference compared to alternatives such as GANs~\cite{heusel2017gans}. This motivates the need for more efficient samplers.


    Recently, much of the progress in accelerating samplers has focused on Ordinary Differential Equation (ODE) solvers~\cite{Lu2022DPMSolverAF,song2020ddim,Wang2023BoostingDM} or on approaches that introduce additional training objectives~\cite{dockhorn2021critically}. A key limitation of ODE-based samplers is their \emph{deterministic} nature: once the initial point is fixed, the generation trajectory is uniquely determined. This restricts their flexibility, whereas stochastic sampling remains crucial in SGMs. Introducing randomness not only enhances sample diversity and fidelity but also enables the computation of likelihoods~\cite{dockhorn2021critically,Karras2022ElucidatingTD}.  

    Stochastic generation via Langevin dynamics is particularly relevant because of its close connection to stochastic gradient descent (SGD)~\cite{Mandt2017StochasticGDlangevin,song2021scorebasedsde}. LD iteratively moves in the direction of the log-density gradient while adding noise, which can be viewed as a noisy gradient ascent procedure that approximates maximum likelihood estimation. This connection suggests that optimization techniques developed for SGD may be fruitfully adapted to improve stochastic sampling in SGMs.  
    
    SGD itself has become a cornerstone for large-scale optimization due to its efficiency and strong empirical success~\cite{Robbins1951}. To accelerate convergence, Polyak’s Heavy Ball (HB) method~\cite{POLYAK19641} augments SGD with a momentum term that captures trends in parameter updates, thereby guiding subsequent iterations more effectively. Its stochastic counterpart, the Stochastic Heavy Ball (SHB) method, introduces a momentum parameter $\beta$ that requires careful tuning. To alleviate this burden, adaptive strategies for automatically adjusting $\beta$ based on historical update statistics have been proposed~\cite{sun2019general, wennormalized}. These methods not only reduce hyperparameter sensitivity but also improve stability and performance, offering a promising direction for momentum-based optimization~\cite{SunHeavyball} and, by extension, for advancing stochastic sampling in SGMs.

\subsection{Contributions}
In this paper, motivated by the limitations of existing approaches and the proven effectiveness of adaptive momentum in SGD, we propose \textbf{Adaptive Momentum Sampling (AMS)} (Algorithm~\ref{Adaptive_Langive_Samp_algo}). After introducing the necessary background, we provide a theoretical analysis of the properties of the sampling distribution induced by AMS and its time evolution. Our main contributions are summarized as follows:

\begin{itemize}
    \item \textbf{A novel plug-and-play sampler.} We introduce a hyperparameter-free sampling method that integrates adaptive momentum from SGD into SGMs. Our sampler operates directly on the reverse SDE process, requires no additional training, and provides a theoretically grounded framework for advancing gradient-based sampling optimization.  

    \item \textbf{Theoretical characterization.} Using the Wasserstein distance, we analyze the sampling process of SGMs within a Markovian framework. We prove that the distribution induced by AMS is stationary and converges to a unique distribution as the number of sampling steps increases. Moreover, we establish explicit relationships between the mean and variance of this distribution and the maximum-likelihood point of $p(\mathbf{x})$, offering new insights into the theoretical underpinnings of SGM sampling.  

    \item \textbf{Empirical validation.} We instantiate AMS as a corrector within Predictor–Corrector (PC) samplers~\cite{song2021scorebasedsde}, referred to as the \textbf{Momentum Corrector (MC)} in Algorithm~\ref{algo::RDP_MC_VE}. Extensive experiments across diverse datasets (images and graphs) and multiple SGMs show that AMS and MC consistently accelerate sampling while improving sample fidelity, demonstrating both strong generality and practical effectiveness.

\end{itemize}

\section{Background}

\subsection{Related Works}

{
Numerous accelerated samplers have been proposed for SGMs. For example, \cite{dockhorn2021critically} introduced Critically-damped Langevin Dynamics (CLD) by incorporating velocity into the conventional Langevin Dynamics framework. However, this approach requires a modified training objective and model re-training. In contrast, \cite{Wang2023BoostingDM,shetty2024momentum,chen2025pid} proposed training-free momentum-based Langevin samplers, which introduce additional hyperparameters that require careful tuning. Our method, by comparison, operates in a plug-and-play manner and does not require any task-specific hyperparameter tuning.
}

Most existing accelerated samplers primarily focus on ODE-based optimization, where deterministic sampling eliminates noise terms, allowing for larger step sizes \cite{Karras2022ElucidatingTD, Lu2022DPMSolverAF, song2020ddim, Wang2023BoostingDM, Wizadwongsa2023DiffusionSWmomentum}. For example, \cite{song2020ddim} introduced Denoising Diffusion Implicit Models (DDIM), a non-Markovian sampler derived from the Euler method. Due to the deterministic nature of non-Markovian sampling, these methods lack diversity in generated images, as no noise is introduced during the sampling process. In contrast, Langevin dynamics inherently incorporates randomness and has a well-established physical interpretation. To the best of our knowledge, there are no existing training-free and hyperparameter-free accelerated Langevin samplers. In this work, we focus on developing stochastic samplers that retain the advantages of Langevin dynamics while achieving efficient sampling acceleration.

\subsection{Score Matching with Langevin Dynamics}
For any continuously differentiable probability density function \(p(\mathbf{x})\), we define its score function as \( \nabla_{\mathbf{x}} \log p(\mathbf{x}) \). Once this score function is available, new samples can be generated from an initial sample using the Langevin dynamics (LD) scheme outlined below 
\begin{equation}\label{eq::langevin_sample}
    \mathbf{x}^{t+1} = \mathbf{x}^{t} +\alpha \nabla_{\mathbf{x}}\log p(\mathbf{x}^{t}) + \sqrt{2\alpha} \boldsymbol{\varepsilon}(\mathbf{x}^{t}),\ \ \ 1\le t\le T,
\end{equation}
where $\boldsymbol{\varepsilon}(\mathbf{x}^{t})$ is a noise term {dependent on or independent of} $\mathbf{x}^{t}$, $\alpha$ is a step size parameter chosen to be sufficiently small, $T$ is chosen to be sufficiently large. In the {annealed Langevin dynamics} \cite{Song2019GenerativeMB}, the noise term is a zero-centered Gaussian noise with covariance $\mathbf{I}$, which can be represented by $\boldsymbol{\varepsilon}$. The convergence of $\mathbf{x}^{t}$ to $p(\mathbf{x})$ is {guaranteed} under certain regularity conditions \cite{roberts1996exponential,welling2011bayesian}. \cite{song2020improved} suggests integrating all the noise information to establish a step size sequence $\{\alpha_i\}_{i=1}^{n}$ with a schedule $\alpha_i = \alpha \sigma_{i}^2/\sigma_{n}^2$, named {Annealed Langevin sampler} (ALS).

The key to LD's \eqref{eq::langevin_sample} application lies in how to estimate the unknown gradient $\nabla \log p(\mathbf{x}^{t})$. Denoising Score Matching (DSM) \cite{Hyv2005ScoreMatching} with Langevin dynamics involves estimating the score function for the objective data distribution $p(\mathbf{x})$. This can be achieved by training a Noise Conditional Score Network (NCSN) \cite{song2020improved}, denoted by $s_{\theta}(\mathbf{x},\sigma)$, by minimizing the following loss \cite{vincent2011connection}
\begin{equation}\label{eq::scorenetwork}
    \frac{1}{2n}\sum_{i=1}^{n}\mathbb{E}_{p_{\rm data}(\mathbf{x})}\mathbb{E}_{p_{\sigma_i}(\tilde{\mathbf{x}}|\mathbf{x})}\Big[  \big\| \sigma_i s_{\theta}(\tilde{\mathbf{x}},\sigma_i) +\frac{\tilde{\mathbf{x}}-\mathbf{x}}{\sigma_i}\big\|^2\Big].
\end{equation}
Here, {$p_{\rm data}(\mathbf{x})$} represents the underlying true data distribution of our target dataset, $\{\sigma_{i}\}_{i=1}^{n}$ are {pre-set noise scales} satisfying {$\sigma_1>\sigma_{2}>\cdots>\sigma_n$, and $p_{\sigma_i}(\tilde{\mathbf{x}}|\mathbf{x})$} represents the corrupted data distribution. In practice, we use the perturbation kernel $p_{\sigma}(\tilde{\mathbf{x}}|\mathbf{x}):=\mathcal{N}(\tilde{\mathbf{x}};\mathbf{x},\sigma^2\mathbf{I})$.

A notable contribution in this area is the establishment of a connection between the forward process and the SDE, along with a solution to the reverse SDE using a reverse diffusion predictor \cite{song2021scorebasedsde}. For the forward process in SMLD, we have 
{
\begin{equation}
    d\mathbf{x} = f(\mathbf{x},t)dt+g(t)d\mathbf{w},
\end{equation}
}
where $\mathbf{w}$ is the standard Wiener process, $f(\cdot, t): \mathbb R^d \rightarrow \mathbb R^d$ is a function {{depicting}} the diffusion process of $\mathbf{x}^{t}$, and {$g(\cdot):\mathbb{R}\rightarrow \mathbb{R}$} is {a scalar function that depends on} the diffusion time $t$. 
We present the corresponding SDE of Variance Exploding (VE) and Variance Preserving (VP) in \cite{song2021scorebasedsde} as follows
\begin{equation}\nonumber
\begin{aligned}
     d\mathbf{x} &= \sqrt{\frac{d[\sigma^2(t)]}{dt}}d\mathbf{w}, \quad (\text{VE}), \\
     d\mathbf{x} &= -\frac{1}{2}\beta(t)\mathbf{x}dt + \sqrt{\beta(t)}d\mathbf{w}, \quad (\text{VP}),
\end{aligned}
\end{equation}
where $\beta(t)$ corresponds to the variation of noise with respect to the time step. Thus, we can deduce the reverse form of the SDE, and the sampling process is conducted along the reverse SDE
{
\begin{equation}\label{eq:reverse_SDE}
    d\mathbf{x} = \big( f(\mathbf{x},t) - g^2(t)\nabla_{\mathbf{x}}\log p_{t}(\mathbf{x})\big)dt + g(t)d\bar{\mathbf{w}}.
\end{equation}
}

\subsection{Integrate Momentum in SMLD}

In SMLD, samples are generated by reversing the process using LD in Eq. \eqref{eq::langevin_sample}. Notably, this update procedure can be interpreted as performing gradient ascent on the log-probability function—using the score \(\nabla_{\mathbf{x}}\log p(\mathbf{x})\) as the ascent direction—which establishes a clear connection between Langevin dynamics and SGD. The key distinction is that LD incorporates stochasticity into the sampling process, whereas traditional SGD operates deterministically.

Consider the following SGD with recursion
\begin{equation}\label{eq::SGD}
    \mathbf{x}^{t+1} = \mathbf{x}^{t} -\alpha\nabla f(\mathbf{x}^t),
\end{equation}
where $\alpha$ denotes the step size hyperparameter, and $f$ is a manually defined loss function. We can regard {the Langevin dynamics} as a special form of SGD with additional noise $\boldsymbol{\varepsilon}$. Thus, by integrating advanced techniques from SGD into Langevin dynamics, we can design an improved sampling schedule for SGMs.

{The Normalized Stochastic Heavy Ball (NSHB) method introduces} a momentum term for SGD Eq. \eqref{eq::SGD} and updates $\mathbf{x}$ with momentum and gradient
{
\begin{equation}\label{eq::NSHBupd}
\mathbf{x}^{t+1}= \mathbf{x}^{t} -\alpha\mathbf{m}^{t}, \mathbf{m}^{t} = \beta_t\mathbf{m}^{t-1} +  (1-\beta_t)\nabla f(\mathbf{x}^{t}),
\end{equation}
} where $\beta \in [0,1)$ is the momentum hyperparameter and requires additional adjustment. 

We treat the gradient \(\nabla f(\mathbf{x}^t)\) as the score function {\(-\nabla_{\mathbf{x}} \log p(\mathbf{x})\)}. By leveraging the trained score network from Eq. \eqref{eq::scorenetwork} and introducing a noise term $\boldsymbol{\varepsilon}$, we can reformulate NSHB. Specifically, we replace the gradient with the trained score network \(s_{\theta}(\mathbf{x}^{t},\sigma)\) to obtain a novel sampling method, as detailed below 
{
\begin{equation}\label{eq::NSHBsamp}
    \mathbf{x}^{t+1} ={\mathbf{x}}^{t} + \tilde{\alpha}_{t} \mathbf{m}^{t} + \sqrt{2\tilde{\alpha}_{t}} \boldsymbol{\varepsilon} (\mathbf{x}^{t}),
\end{equation}
}
where $\mathbf{m}^{t} = \beta_{t} \mathbf{m}^{t-1} + (1-\beta_t) s_{\theta} (\mathbf{x}^{t},\sigma_t)$.

\section{Technical Tools}

    \begin{figure*}[!ht]
        \centering
        
        \subfloat[Local-$\mu$ along SDE Trajectory]{\includegraphics[width=0.45\linewidth,height=0.23\textwidth]{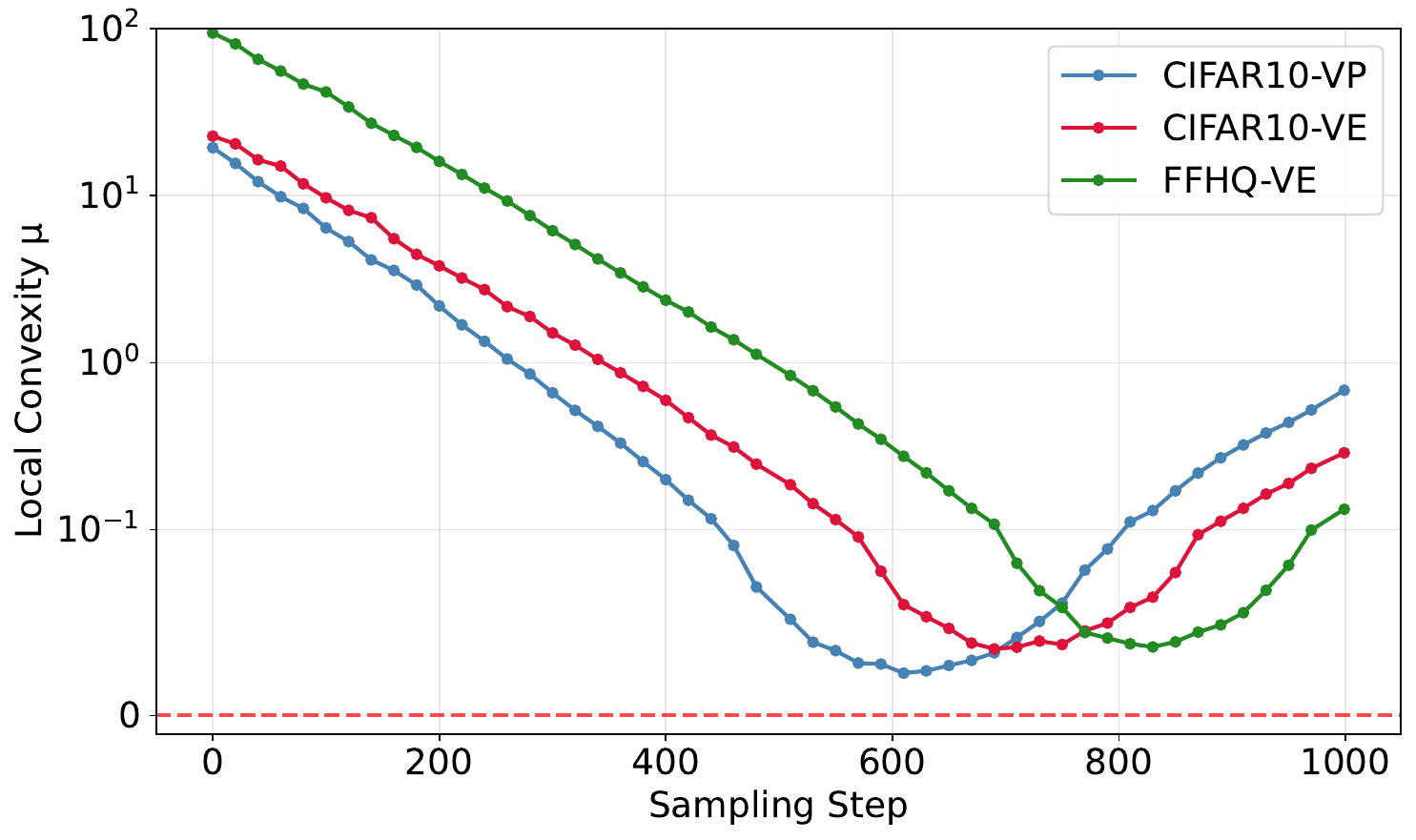}}
        \hfill
        \subfloat[Local Condition Number Analysis]{\includegraphics[width=0.45\linewidth,height=0.23\textwidth]{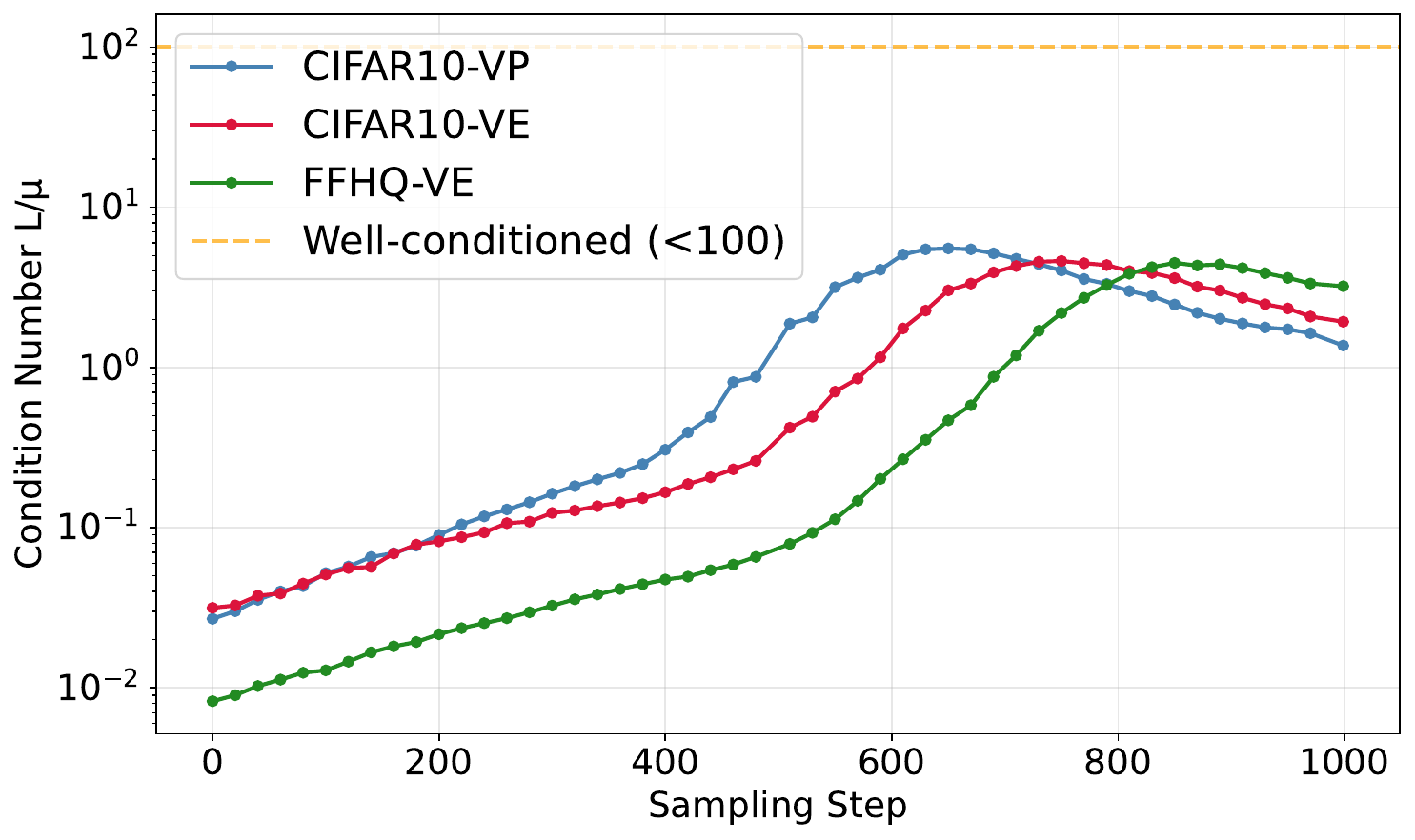}}
        
        \caption{{
        \textbf{Experiments on the strong convexity of SGM sampling trajectories.} Left: The local convexity parameter $\mu$ as a function of the sampling steps, where $\mu$ remains greater than 0 throughout the entire sampling trajectory. Right: Analysis of the condition number $L/\mu$ along the sampling trajectory, where the overall condition number remains below 100, indicating a well-behaved sampling process.
        }} 
     \label{fig:convexity}
    \end{figure*}

\subsection{Notation}

{To begin our analysis, we introduce some notation to improve readability.} Consider the finite-dimensional vector space $\mathbb R^d$ {equipped with the canonical inner product } $<\cdot,\cdot>$. We use lowercase letters and lowercase boldface letters to separately denote scalars and vectors like  $\mathbf{x}=(x_1,x_2,\cdots,x_d)$, $\|\mathbf{x}\|$ denotes its $\ell_2$ norm. The symbol $\otimes$ denotes the Kronecker product.  Uppercase boldface letters denote a matrix, i.e., $\mathbf{I}$ denotes the identity matrix. For $a =\mathcal{O}(b)$, we specify that there exists a positive constant $0<C<+\infty$, such that $a<Cb$. $\mathcal{B}(\mathbb{R}^d)$ stands for the Borel-$\sigma$ field of $\mathbb{R}^d$. We use $\mathcal{P}(\mathbb{R}^d)$ to represent the space consisting of all Borel {probability measures} $\theta$ which satisfy $\int_{{\mathbb{R}^d}} \|\mathbf{z}\|^2\theta(d\mathbf{z})<+\infty$. We denote the derivative operation and Hessian matrix for a function $f({\mathbf{x}}): \ \mathbb{R}^d \rightarrow \mathbb{R}$ by $ \nabla f({\mathbf{x}})$ and $\nabla^2 f({\mathbf{x}})$,   respectively.

For probability measures $\theta,\hat{\theta}$ in $\mathcal{P}(\mathbb{R}^d)$, define the Wasserstein distance of order 2
between $\theta$ and $\hat{\theta}$ as
{
\begin{equation}\nonumber
    \mathcal{W}_{2}(\theta,\hat{\theta}) := \underset{\xi\in\Pi(\theta,\hat{\theta})}{\inf} (\int\|\mathbf{x}-\mathbf{y}\|^2 \xi(d\mathbf{x},d\mathbf{y}))^{\frac{1}{2}},
\end{equation}
}
where $\Pi(\theta,\hat{\theta})$ is a set of all joint probability distributions between $\theta$ and $\hat{\theta}$.

We collect the assumption underlying our analysis. Let $f : \mathbb R^{d} \rightarrow \mathbb R$ {satisfy} the following:

\begin{assumption}[$L$-Lipschitz] \label{assump:lipschitz}
    The function $f$ is differentiable, and the gradient is $L$-Lipschitz, i.e., for all $\mathbf{x}, \mathbf{y} \in\mathbb R^d$
    \begin{equation}\nonumber
\|\nabla f(\mathbf{x})-\nabla f(\mathbf{y})\| \le L\|\mathbf{x}-\mathbf{y}\|.
\end{equation}
\end{assumption}

\begin{assumption}[Strong convexity] \label{assumption_mu} The function $f$ is strongly convex with a constant $\mu >0$, i.e., for all       $\mathbf{x},\mathbf{y} \in \mathbb R^{d}$, and $t\in[0,1) $ such that
        \begin{equation}\nonumber
        f(t\mathbf{x}+(1-t)\mathbf{y}) \le tf(\mathbf{x}) + (1-t)f(\mathbf{y}) - \frac{\mu}{2}t(1-t)\|\mathbf{x}-\mathbf{y}\|^2.
        \end{equation}
\end{assumption}
\subsection{Markov Chain Associated with Momentum}
In this section, we represent NSHB in Eq. \eqref{eq::NSHBupd} within the framework of Markov chains and introduce key concepts from Markov theory \cite{ccinlar2011probability, meyn2012markov}. We define the variable \(\mathbf{z}^{t+1} \in \mathbb{R}^{2d}\) as 
$ \mathbf{z}^{t+1} = \begin{bmatrix} \mathbf{x}^{t+1}, \mathbf{x}^{t} \end{bmatrix}^{T},$ and use $\mathbf{x}^{*}$ to denote the point with the maximum likelihood solution for the distribution $p(\mathbf{x})$ and $\mathbf{z}^{*} = \begin{bmatrix}
    \mathbf{x}^{*}, \mathbf{x}^{*}
\end{bmatrix}^{T}$. From this, we derive the Markov kernel \(R_{\alpha}\) on \((\mathbb{R}^{2d}, \mathcal{B}(\mathbb{R}^{2d}))\) associated with the NSHB iterates in Eq. \eqref{eq::NSHBupd} by $\{\mathbf{z}^t\}_{t\geq 1}$, where $\alpha$ denotes the step size
$$
R_{\alpha}(\mathbf{z}^{t}, C) = P(\mathbf{z}^{t+1} \in C \mid \mathbf{z}^{t}).
$$
For any integer $t> 1$, we can define the Markov kernel $R_{\alpha}^{t}$ recursively by $R_{\alpha}^{1} = R_{\alpha}$, and
$$
    R_{\alpha}^{t+1} (\hat{\mathbf{v}},C):= \underset{\mathbb R^{2d}}{\int}R_{\alpha}^{t}(\hat{\mathbf{v}},d\mathbf{v})R_{\alpha}(\mathbf{v},C),
$$ which can be rewritten as $ P(\mathbf{z}^{t+1} \in C \mid \mathbf{z}^{0}= \hat{\mathbf{v}}) $.

\section{Algorithm and Mathematical Analysis}
In this section, we first introduce a method for computing adaptive momentum and present a momentum-based sampling algorithm (detailed in Algorithm \ref{Adaptive_Langive_Samp_algo}). Next, we provide a theoretical analysis of the proposed method, demonstrating that under specific conditions, the data distribution—characterized as a function of the sampling time step $ t $—is both stationary and unique, \textbf{aligning with the maximum likelihood of the gradients learned by the model}. Additionally, we examine the properties of the mean and variance of this distribution.
\subsection{Our Algorithm}
{
The hyperparameter \(\beta\) plays a crucial role in the success of the sampling process and may require careful tuning. Prior work has shown that, under mild assumptions, the optimal choice of \(\beta\) in Eq. \eqref{eq::NSHBupd} for quadratic optimization is given by \(\beta = \left(\frac{1-\alpha \mu}{1+\alpha \mu}\right)^2\) \cite{wennormalized} (see Lemma~\ref{lemma:best_beta_lemma} in Appendix). However, this formulation depends on \(\mu\), which is typically unknown in practice. To overcome this limitation, \cite{wennormalized} proposed an approximation scheme, 
\begin{equation}
    \nonumber
    \underset{k\rightarrow \infty}{\lim}\frac{\|{\mathbf{g}}^k-{\mathbf{g}}^{k-1}\|}{\|{\mathbf{x}}^{k}-{\mathbf{x}}^{k-1}\|}=\mu,
\end{equation}
to estimate the optimal $\beta$, where $\mathbf{g}^t$ denotes the gradient of the loss function evaluated at $\mathbf{x}^t$, {thereby reducing the need for manual parameter tuning through the following update rule:}
\begin{small}
    \begin{equation}\label{eq::The_best_beta}
    \beta_{t+1} =\left\{
    \begin{array}{lr}
          {\rm Proj}_{[0,1-\delta]} \left(\left(\dfrac{1-\alpha\dfrac{\|{\mathbf{g}}^t-{\mathbf{g}}^{t-1}\|}{\|{\mathbf{x}}^t-{\mathbf{x}}^{t-1}\|}}{1+\alpha\dfrac{\|{\mathbf{g}}^t-{\mathbf{g}}^{t-1}\|}{\|{\mathbf{x}}^t-{\mathbf{x}}^{t-1}\|}}\right)^2\right)  &t\ge 1,\\
          0&t=0,
    \end{array}
    \right.
\end{equation}
\end{small}
where ${\rm Proj}_{[0,1-\delta]}(\cdot):=\max(0, \min(\cdot, 1-\delta))$ with a threshold $\delta$ that limits $\beta_{t} \in [0,1-\delta]$.}

We follow the same strategy as LD for the step size $\alpha$ used in our approach. The momentum incorporation introduces noise from the previous step, scaled by a factor of $\beta$. To ensure that the signal-to-noise ratio remains unchanged, we scale $\alpha$ in LD by a factor of $(1+\beta)^2$, resulting in $\tilde{\alpha}_{t+1} = \alpha_{t+1} (1+\beta_{t})^2$. In summary, we have developed a novel sampling method for SGMs that demonstrates robust Markov chain convergence properties. Detailed theoretical guarantees are provided in the Appendix (see Lemmas \ref{lemma:best_beta_lemma} and \ref{lemma:approximate_mu}). This method ensures convergence under specific assumptions and exhibits strong stochastic characteristics. Its pseudo-code is presented in Algorithm \ref{Adaptive_Langive_Samp_algo}, and we refer to the method as Adaptive Momentum Sampling (AMS). Additionally, we present pseudo-code for our Momentum-Corrector (MC) combined with the Reverse Diffusion (RD) VE-SDE solver \cite{song2021scorebasedsde} in Algorithm \ref{algo::RDP_MC_VE}.

\begin{algorithm}[!ht]
	\caption{Adaptive Momentum Sampling (ours)}
    \begin{algorithmic}
    \REQUIRE {step size $\alpha>0$, margin parameter $\delta>0$, score-net $s_{\theta}$, noise scale hyperparameters {$(\sigma_i)_{i=1}^{n+1}$}, $n_{\sigma}$}

    Initialize $\mathbf{x}, \mathbf{m} =\mathbf{0}, \beta = 0$
    \FOR{$i \leftarrow 1$ to $n$}
        \STATE $\alpha_{i} = \alpha \sigma_{i}^{2}/{\sigma_{n}^{2}}, \tilde{\alpha} = \alpha_{i} (1 + \beta)^2$ \COMMENT{calculate step size}
         \FOR{$k \leftarrow 1$ to $n_{\sigma}$ }
            \STATE Draw Gaussian noise $\boldsymbol{\varepsilon} \sim \mathcal{N}(\mathbf{0},\mathbf{I})$
            \STATE Calculate $\beta$ and update $\mathbf{x}$ by Eqs. \eqref{eq::The_best_beta}, and \eqref{eq::NSHBsamp}
         \ENDFOR
    \ENDFOR
        \IF{Denoised}
    \STATE return $\mathbf{x}+\sigma_{n}^{2}s_{\theta}(\mathbf{x},\sigma_n)$ \COMMENT{additional denoise step}
    \ELSE  
    \STATE return $\mathbf{x}$
    \ENDIF
    \end{algorithmic}\label{Adaptive_Langive_Samp_algo}
\end{algorithm}

{Note that Algorithms \ref{Adaptive_Langive_Samp_algo} and \ref{algo::RDP_MC_VE} introduce an additional hyperparameter, $\delta$. This parameter constrains the adaptively computed momentum $\beta_t$, ensuring that it remains strictly below one in extreme cases. In the subsequent experiments, we demonstrate that our approach is effectively hyperparameter-free. Specifically, the performance of our method is insensitive to the choice of $\delta$, and the adaptive momentum achieves strong performance without requiring task- or model-specific tuning.}


\begin{figure*}[!t]
    \centering
    \rotatebox{90}{~~~~~~~~~~~~FID}
    \subfloat[RD-LC sampler, 150 NFE]{\includegraphics[width=0.32\linewidth,height=0.2\textwidth]{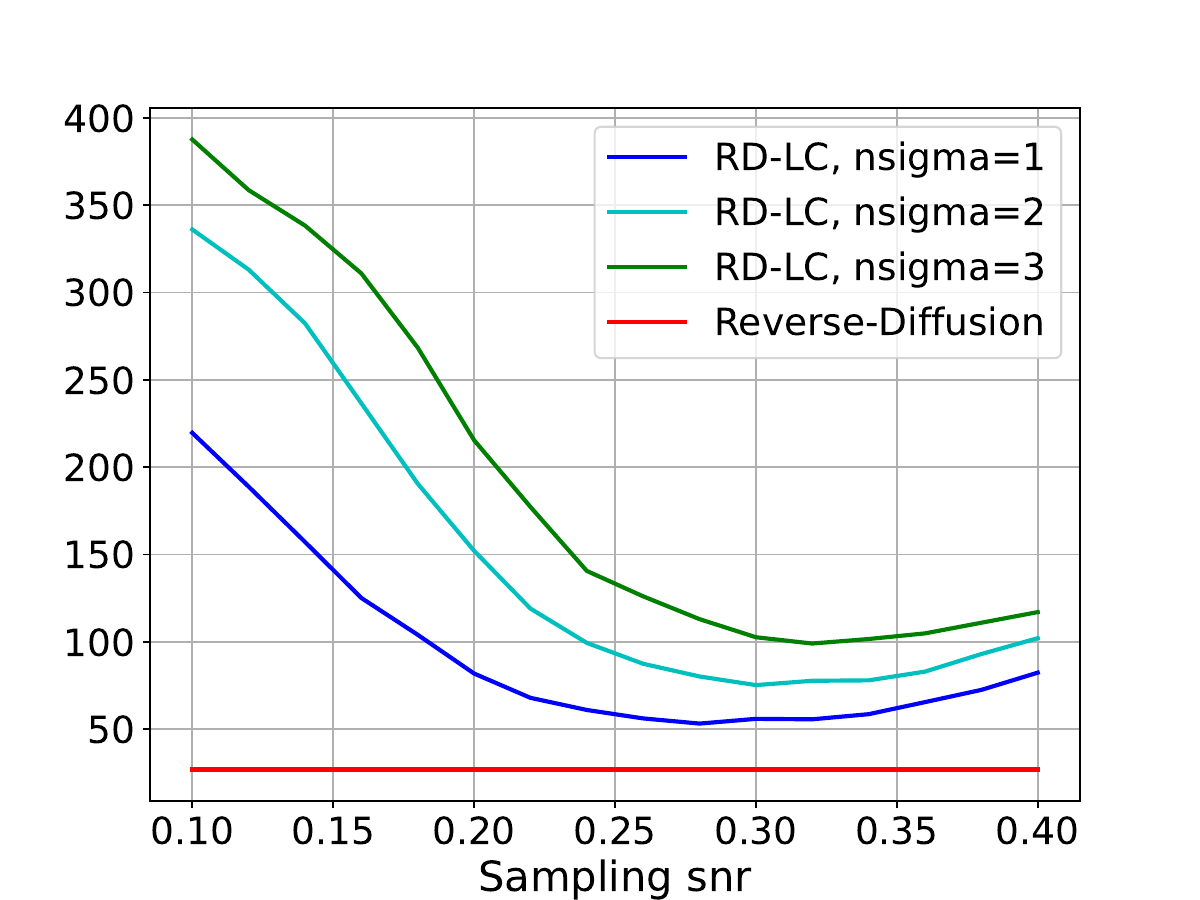}}
    \hfil
    \subfloat[EM-MC sampler, 150 NFE]{\includegraphics[width=0.32\linewidth,height=0.2\textwidth]{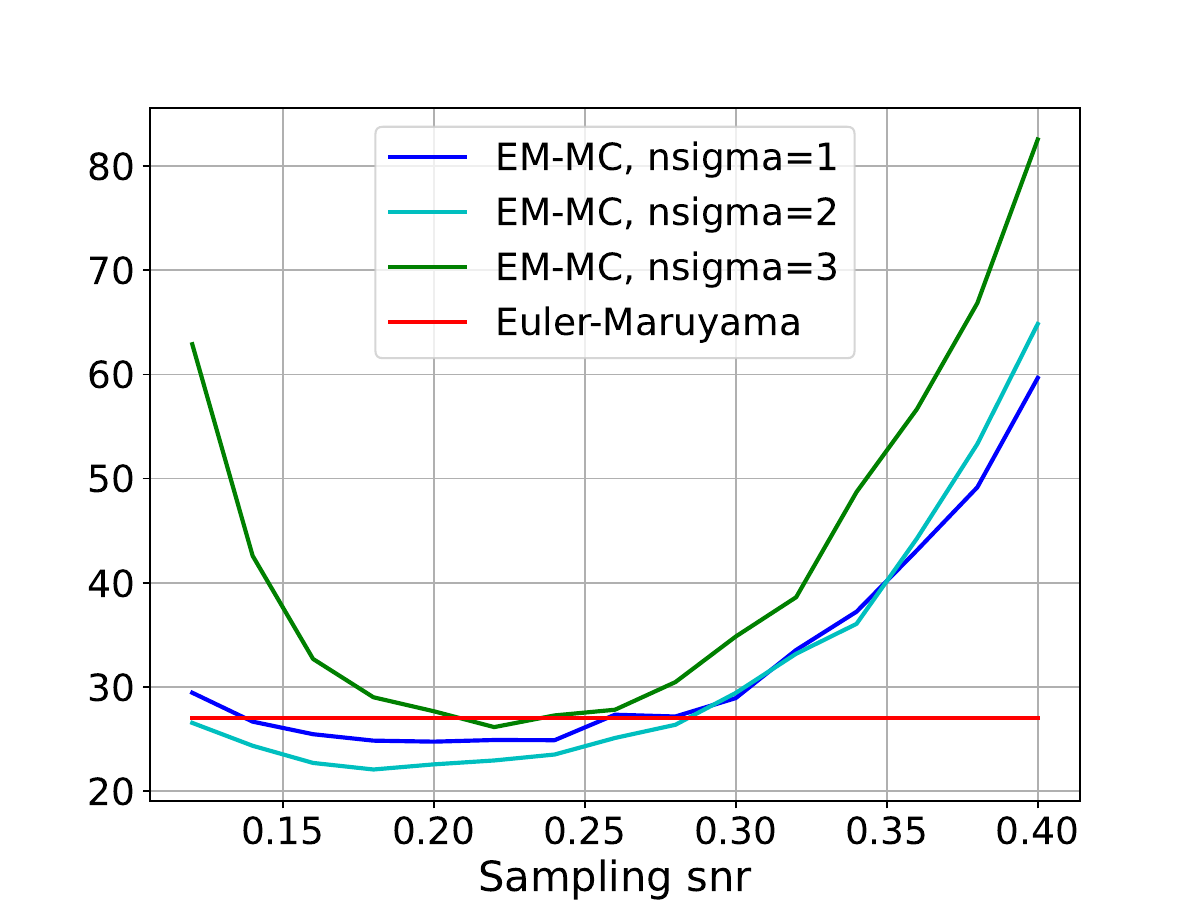}}
    \hfil
    \subfloat[RD-MC sampler, 150 NFE]{\includegraphics[width=0.32\linewidth,height=0.2\textwidth]{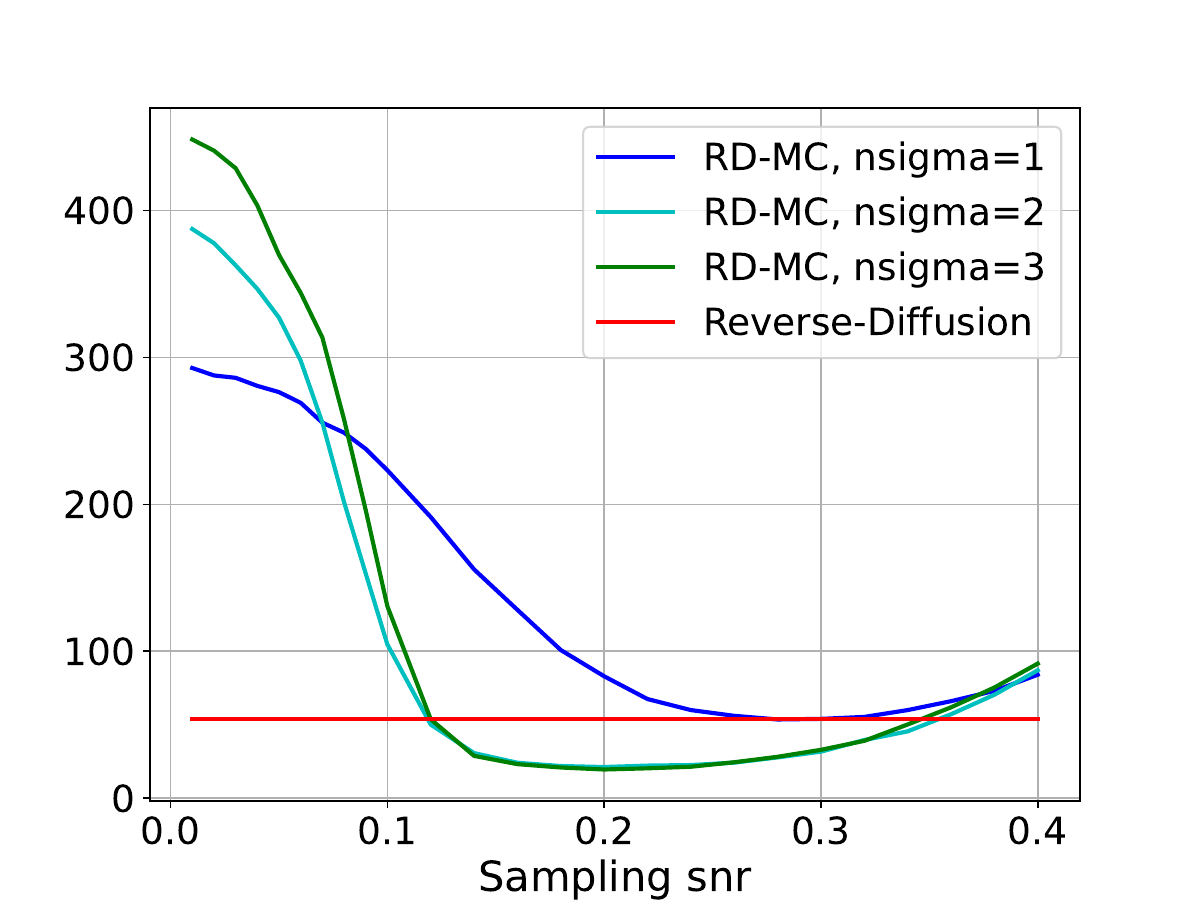}}
    \\
    \rotatebox{90}{~~~~~~~~~~~~FID}
    \subfloat[RD-LC sampler, 270 NFE]{\includegraphics[width=0.32\linewidth,height=0.2\textwidth]{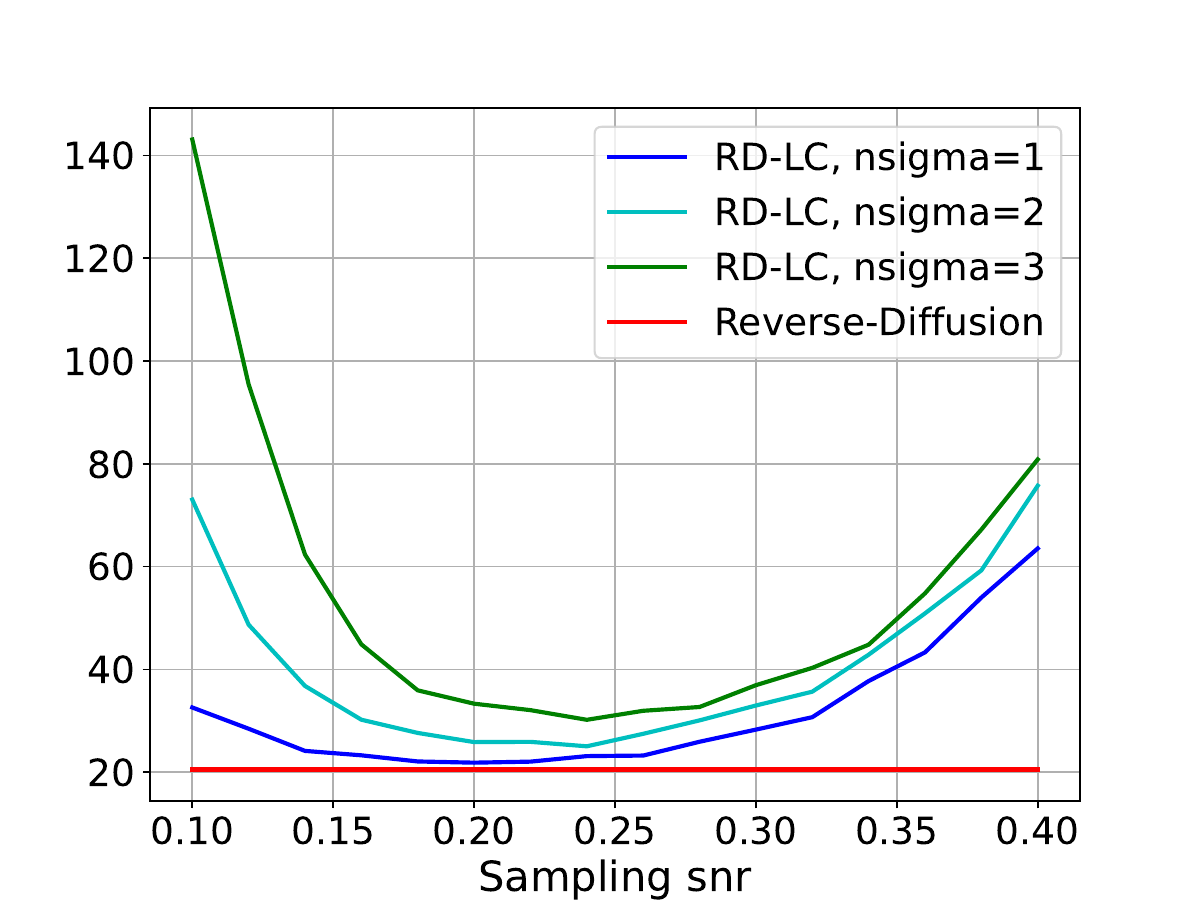}}
    \hfil
    \subfloat[EM-MC sampler, 270 NFE]{\includegraphics[width=0.32\linewidth,height=0.2\textwidth]{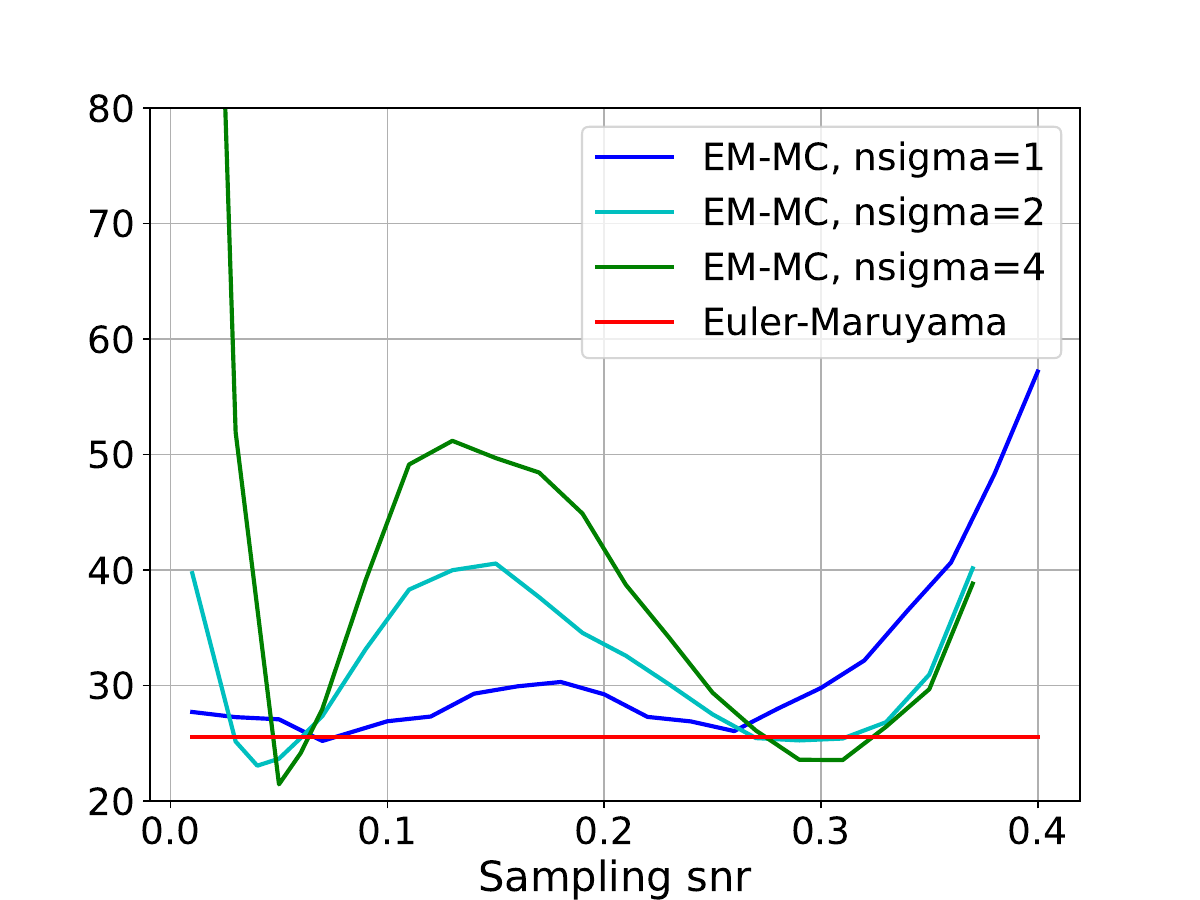}}
    \hfil
    \subfloat[RD-MC sampler, 270 NFE]{\includegraphics[width=0.32\linewidth,height=0.2\textwidth]{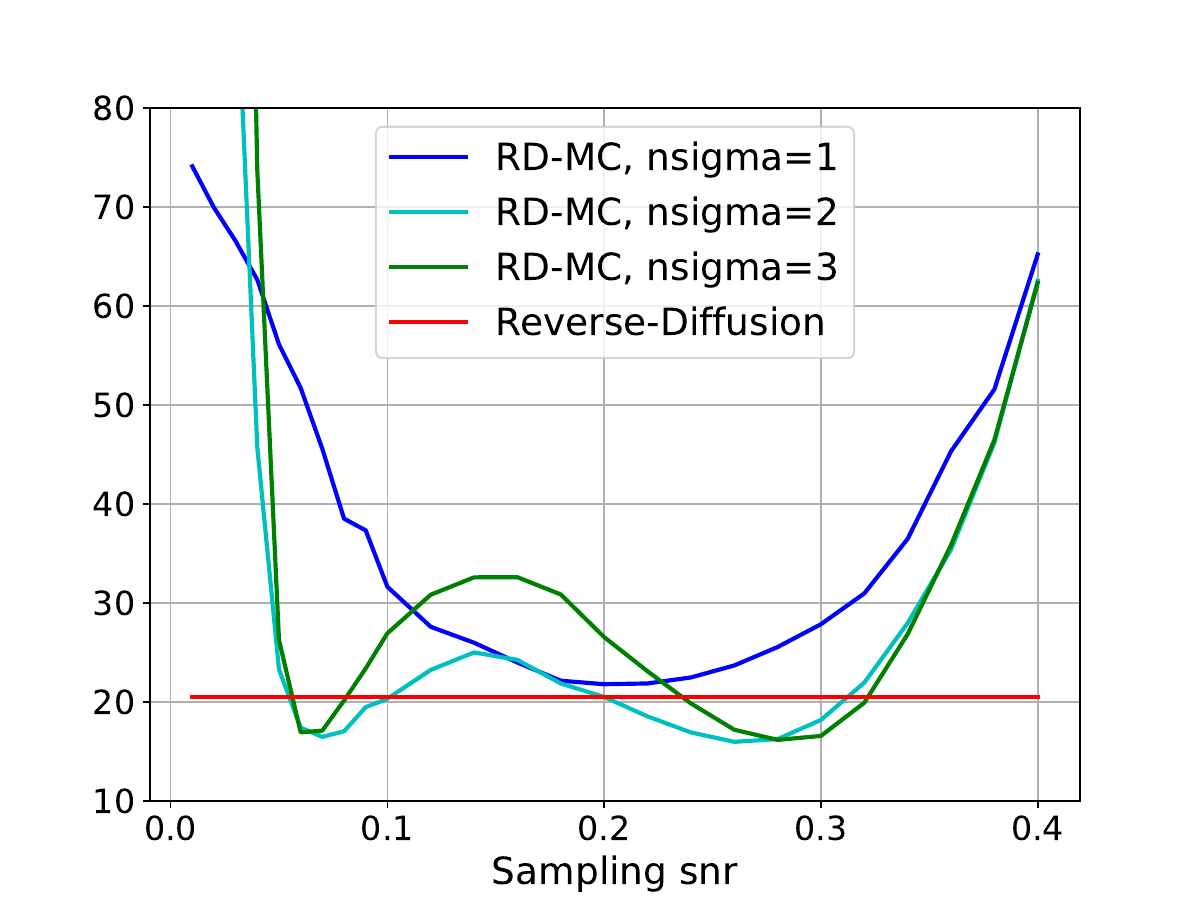}}
    \caption{\textbf{Partial estimate of FID $\downarrow$ (2k samples) with different initial $\epsilon$.} {
    We observe that MC (ours) can improve image quality with an appropriate initial Signal-to-Noise Ratio (SNR) hyperparameter $\alpha$.
    }} 
 \label{fig:150_NFE_nsigma_snr}
\end{figure*}

\subsection{Quadratic Case}
In this section, we investigate the efficiency property when $f$ takes the form of a quadratic function: 
{
\begin{equation}\nonumber
    f(\mathbf{x})=\frac{1}{2}\mathbf{x}^T\mathbf{A}\mathbf{x}+\mathbf{b}^T\mathbf{x}+c,\ \nabla f(\mathbf{x}) = \mathbf{A}\mathbf{x}+\mathbf{b}.
\end{equation} 
}
Furthermore, we denote the smallest and largest eigenvalues of $\mathbf{A}$ as $\mu>0$ and $L$ respectively.  To simplify the expression, we assume $\beta$ remains constant.  Hence, we can reformulate Eq. \eqref{eq::NSHBupd} with an additional noise term as
{
\begin{equation}\label{markovz}
    \mathbf{z}^{t+1}
    =
    \mathbf{T}\mathbf{z}^{t}
    +
    \begin{bmatrix}
    -\alpha(1-\beta)\mathbf{b}\\
    \mathbf{0}
    \end{bmatrix}
    +
    \boldsymbol{\varepsilon}^{t+1},
    \end{equation}
}
where \(\{\boldsymbol{\varepsilon}^{t}\}_{t\ge 1}\) is an i.i.d.\ centered additive noise sequence independent of the chain, and we denote its covariance by
\begin{equation}\nonumber
    \boldsymbol{\Sigma}_{\boldsymbol{\varepsilon}}:=\mathbb{E}\left[\boldsymbol{\varepsilon}^{t}(\boldsymbol{\varepsilon}^{t})^{T}\right].
\end{equation}
\begin{algorithm}[!ht]
    \caption{RD-MC sampler for VE-SDE}
    \begin{algorithmic}
    \REQUIRE hyperparameters $(\epsilon_{i})_{i=1}^{n+1}>0$, $\delta>0$, score-net $s_{\theta}$, noise scale $(\sigma_i)_{i=1}^{n+1}$, $n_{\sigma}$.
    Initialize $\mathbf{x}, \mathbf{m}^{0}=\mathbf{0},\beta=0$.
    \FOR{$i \leftarrow 1$ to $n$}
        
        \STATE Draw Gaussian noise $\boldsymbol{\varepsilon} \sim \mathcal{N}(\mathbf{0},\mathbf{I}).$
        \STATE $\mathbf{x} \leftarrow \mathbf{x} + (\sigma_{i+1}^2-\sigma_{i}^2)s_{\theta}(\mathbf{x},\sigma_{i+1}).$
        \STATE $\mathbf{x} \leftarrow \mathbf{x} + \sqrt{\sigma_{i}^2-\sigma_{i+1}^2}{\boldsymbol{\varepsilon}}.$
         \FOR{$k \leftarrow 1$ to $n_{\sigma}$ }
            \STATE Draw Gaussian noise $\boldsymbol{\xi} \sim \mathcal{N}(\mathbf{0},\mathbf{I}).$
            \STATE $\alpha_i = 2 \big(\epsilon_i (1 + \beta) ^ 2\| \mathbf{m}^{i}\|_2 / \| \boldsymbol{\xi} \|_2\big )^{2}.$
            \STATE Calculate $\beta$ and update $\mathbf{x}$ by Eqs. \eqref{eq::The_best_beta}, and \eqref{eq::NSHBsamp}.
         \ENDFOR
    \ENDFOR
        \IF{Denoised}
    \STATE return $\mathbf{x}+\sigma_{n}^{2}s_{\theta}(\mathbf{x},\sigma_n).$
    \ELSE  
    \STATE return $\mathbf{x}.$
    \ENDIF
    \end{algorithmic}\label{algo::RDP_MC_VE}
\end{algorithm}
where we define \begin{equation}\nonumber
    \mathbf{T}:=\begin{bmatrix}
        (1+\beta)\mathbf{I}-\alpha(1-\beta)\mathbf{A} &-\beta \mathbf{I} \\
        \mathbf{I} & \mathbf{0}
    \end{bmatrix}.
\end{equation}

After presenting the Markov chain definitions, we provide the convergence results and properties of the sample distribution obtained by our sampling method in Algorithm \ref{Adaptive_Langive_Samp_algo}.

\begin{proposition}\label{proposition_pi_stationary}
{
Suppose $\alpha \in(0,1/L]$ and $\beta$ is chosen such that $\rho(\mathbf{T})<1$, as verified by Eq.~\eqref{lambda_max_T}. Then the Markov chain $(\mathbf{z}^t)_{t\ge 0}$ defined by the recursion Eq.~\eqref{markovz} admits a unique stationary distribution $\pi_{\alpha}^*$ such that $\pi_{\alpha}^* R_{\alpha} = \pi_{\alpha}^*$. Additionally, for any \(0<\epsilon<1-\rho(\mathbf{T})\), there exists \(C_\epsilon>0\) such that for all $\hat{\mathbf{z}}\in\mathbb R^{2d}$ and $t\in \mathbb N^*$,
\[
    \mathcal{W}_{2}(R_{\alpha}^{t}(\hat{\mathbf{z}},\cdot), \pi_{\alpha}^*) 
    \leq 
    C_\epsilon(\rho(\mathbf{T})+\epsilon)^{t}
    \left(\int_{\mathbb R^{2d}}\|\mathbf{z}-\hat{\mathbf{z}}\|^2\pi_{\alpha}^*(d\mathbf{z})\right)^{1/2}.
\]
Here, \(\pi_{\alpha}^{*}\) is the stationary distribution of the augmented Markov chain \(\mathbf{z}^{t}=(\mathbf{x}^{t},\mathbf{x}^{t-1})\) on \(\mathbb{R}^{2d}\), whose two marginals are denoted by \(\pi_{\alpha}\). Moreover, $\rho(\mathbf{T}) $ represents the spectral radius of $ \mathbf{T} $, $\epsilon$ arises from Gelfand's theory, and we can demonstrate that $ \rho(\mathbf{T}) < 1 $ under constrained $ \alpha $ and $ \beta $ (see the Appendix).}
\end{proposition}

In Proposition \ref{proposition_pi_stationary}, we consider $\hat{\mathbf{z}}$ as the initialization of the sampling process. This proposition establishes that the iterates in Eq. \eqref{markovz} converge to a unique stationary distribution. Consequently, we analyze the properties of the Markov chain starting from $\mathbf{z}^{0}$, which follows the distribution $ \pi_{\alpha}^*$.

\begin{proposition} \label{proposition_expec_quard}
Assume $\mathbf{A}$ is a positive definite matrix and $\alpha,\beta$ are chosen such that $\rho(\mathbf{T})<1$ as in Eq.~\eqref{lambda_max_T}. Then $\mathbf{I}-\mathbf{T}\otimes\mathbf{T}$ is invertible. Moreover, if $\pi_{\alpha}^{*}$ is the stationary distribution of Eq. \eqref{markovz}, then
    \begin{equation}\nonumber
        \begin{aligned}
            &\operatorname{vec}\left(\underset{\mathbb R^{2d}}{\int} (\mathbf{z}-\mathbf{z}^*)^{\otimes 2}\pi_{\alpha}^*(d\mathbf{z})\right) \\
            &= (\mathbf{I}-\mathbf{T}\otimes\mathbf{T})^{-1}\operatorname{vec}(\boldsymbol{\Sigma}_{\boldsymbol{\varepsilon}}).
        \end{aligned}
    \end{equation}
\end{proposition}
where $\boldsymbol{\Sigma}_{\boldsymbol{\varepsilon}}=\mathbb{E}[\boldsymbol{\varepsilon}^{t}(\boldsymbol{\varepsilon}^{t})^{T}]$ is the covariance matrix of the additive noise, and $\mathbf{z}^{*} = \begin{bmatrix}
    \mathbf{x}^{*}, \mathbf{x}^{*}
\end{bmatrix}^{T}$. These two propositions highlight the fact that, for a given quadratic function, the sample distribution is stationary and unique.

\subsection{Strong Convexity Case}
In this section, we dive into analyzing the convergence of Eq. \eqref{eq::NSHBupd} in the strongly convex case. {In the local convergence analysis, we consider the terminal annealed regime where
\[
\sigma_\alpha^2=\alpha\kappa(\mathbf{x}^*)+\mathcal{O}(\alpha^{3/2}).
\]
Thus, the local Langevin perturbation has covariance
\[
2\tilde{\alpha}\sigma_\alpha^2 \mathbf{I}
=
\alpha^2\mathcal{C}(\mathbf{x}^*)+\mathcal{O}(\alpha^{5/2}),
\]
where \(\mathcal{C}(\mathbf{x}^*)=2(1+\beta)^2\kappa(\mathbf{x}^*)\mathbf{I}\).
}
\begin{proposition}\label{proposition_conv}
Assume $\alpha \in (0,1/L)$ and $f$ is strongly convex. Let
\begin{equation}\nonumber
    \mathbf{P}:=\begin{bmatrix}\mathbf{I}&\mathbf{0}\end{bmatrix}, 
    \mathbf{x}-\mathbf{x}^{*}=\mathbf{P}(\mathbf{z}-\mathbf{z}^{*}), \mathbf{H}:=\nabla^2 f(\mathbf{x}^*),
\end{equation}
{and define the local transition matrix
\[
\mathbf{T}_{\mathbf{H}}:=
\begin{bmatrix}
(1+\beta)\mathbf{I}-\alpha(1-\beta) \mathbf{H} & -\beta\mathbf{I}\\
\mathbf{I} & \mathbf{0}
\end{bmatrix}.
\]} Suppose the local additive noise in the augmented recursion satisfies
\begin{equation}\nonumber
    \boldsymbol{\Sigma}_{\boldsymbol{\varepsilon}}
    :=\mathbb{E}\left[\boldsymbol{\varepsilon}_{z,t}\boldsymbol{\varepsilon}_{z,t}^{T}\right]
    =
    \begin{bmatrix}
    \alpha^2\mathcal{C}(\mathbf{x}^{*})+\mathcal{O}(\alpha^{\frac{5}{2}}) & \mathbf{0}\\
    \mathbf{0} & \mathbf{0}
    \end{bmatrix},
\end{equation}
where $\boldsymbol{\varepsilon}_{z,t}=\begin{bmatrix}\boldsymbol{\varepsilon}_{\mathbf{x},t}\\ \mathbf{0}\end{bmatrix}$ is the augmented additive noise. If \(\rho(\mathbf{T}_{\mathbf{H}})<1\), then

\small{\begin{equation}\nonumber
\begin{aligned}
    &\operatorname{vec}\left(\underset{\mathbb R^d}{\int} 
    (\mathbf{x}-\mathbf{x}^*)^{\otimes 2}\pi_{\alpha}(d\mathbf{x})\right) \\
    &\quad =
    (\mathbf{P}\otimes\mathbf{P})(\mathbf{I}-\mathbf{T}_{\mathbf{H}}\otimes\mathbf{T}_{\mathbf{H}})^{-1}
    \operatorname{vec}\left(
    \begin{bmatrix}
    \alpha^2\mathcal{C}(\mathbf{x}^{*}) & \mathbf{0}\\
    \mathbf{0} & \mathbf{0}
    \end{bmatrix}
    \right)
    +\mathcal{O}(\alpha^{\frac{3}{2}}).
\end{aligned}
\end{equation}}
\end{proposition}
Our theoretical analyses demonstrate that our method ensures a stationary and unique sampling distribution and tends to sample high probability density points with high frequency. Furthermore, the converged distribution remains closely aligned with the distribution learned by the SGM, reinforcing the theoretical soundness of our approach.

{
\textbf{On the Validity of Assumptions.} Our convergence analysis is built under classical assumptions such as smoothness and strong convexity, which provide theoretical tractability and help interpret the behavior of the proposed sampler. Although such assumptions may not strictly hold in general SGMs, they are approximately satisfied in local regions of the loss landscape, especially when the model is close to convergence. Moreover, in practice, SGMs often include regularization (e.g., weight decay, score matching penalties) that induces a locally convex structure. To verify this local strong convexity, we perform 1000-step SDE sampling to generate CIFAR-10 and FFHQ images using the pre-trained VE/VP model~\cite{Karras2022ElucidatingTD}. We estimate the local convexity coefficient $\mu$, and the {condition number $L/\mu$} of the Hessian. Specifically, at each trajectory point $\mathbf{x}^{t}$ with noise level $\sigma_t$, we define the residual function {for a pre-trained SGM as follows}
    \begin{equation}\nonumber
    f(\mathbf{x}; \sigma_t) = \frac{1}{2}\|s_{\theta}(\mathbf{x}, \sigma_t) - s_{\theta}(\mathbf{x}^t, \sigma_t)\|^2,
    \end{equation}
    where $s_{\theta}$ denotes the trained score-based network and $s_{\theta}(\mathbf{x}^t, \sigma_t)$ serves as the local reference output for evaluating quadratic growth. This local quadratic loss measures the deviation of the score network output around $\mathbf{x}$, enabling us to probe the local convexity of the score landscape via its second-order geometry.}

{
The strong convexity coefficient $\mu$ is computed via the quadratic growth condition: for a random perturbation $\mathbf{d}$ with radius $r$, we have
    \begin{equation}\nonumber
    \mu = \frac{2\left[f(\mathbf{x}^{t}+\mathbf{d}) - f(\mathbf{x}^{t}) - \langle\nabla f(\mathbf{x}^{t}),\mathbf{d}\rangle\right]}{\|\mathbf{d}\|^2}.
    \end{equation}
    The Lipschitz constant $L$ is estimated via the following perturbation
    \begin{equation}
        \nonumber
        L = \frac{\|s_{\theta}(\mathbf{x} + \boldsymbol{\varepsilon}, \sigma_t) - s_{\theta}(\mathbf{x}, \sigma_t) \|}{\| \boldsymbol{\varepsilon} \|}.
    \end{equation}
    We sample $n=50$ random directions at each trajectory point with radius $r \in \{0.01, 0.02, 0.05, 0.1\}$ and aggregate the resulting statistics. The local condition number is computed from $L/\mu$. While these estimates are approximate due to finite sampling and local perturbations, they provide valuable insights into the local geometry of the score-based generative model. The results are shown in Figure~\ref{fig:convexity}. The results demonstrate that throughout the sampling process, the local convexity coefficient $\mu$ remains positive, and the condition number remains well-behaved. This provides experimental evidence supporting the validity of our strong convexity assumption.
}

\section{Numerical Results}\label{sec::numerical results}

\begin{table}[!ht]
    \centering
    \caption{FID $\downarrow$ scores and sampling times (500 images) of different samplers on NCSN (VE) across different $n_\sigma$.} \label{tab:ALS_nsigma}
    \renewcommand{\arraystretch}{1.1}
    \setlength{\tabcolsep}{10pt}
    \begin{tabular}{lcccc}
    \toprule
         Sampler & NFE & 50 & 150 & 250\\
         \midrule
        \multirow{2}{*}{\textbf{ALS}} & $n_\sigma = 1$ & $57.71$& $17.16$& $12.29$\\
        & $n_\sigma = 5$ & $42.74$& $13.87$& $10.98$\\
        \midrule
        \multicolumn{2}{l}{Sampling time} & 18.64s & 48.47s & 79.03s \\
        \midrule
        \multirow{2}{*}{\textbf{AMS}} & $n_\sigma = 1$ & $30.52$& $17.16$& $12.29$\\
        & $n_\sigma = 5$ & $26.37$& $11.26$& $10.08$\\
        \midrule
        \multicolumn{2}{l}{Sampling time} & 18.69s & 48.50s & 79.33s \\
        \bottomrule
    \end{tabular}
\end{table}

\subsection{Experimental Setup}

{
To evaluate the performance and applicability of our approach, we conduct experiments on both image and graph synthesis tasks across a variety of datasets and model architectures. We begin by outlining the key experimental details.
}
\textbf{Datasets and Models.} 
For image generation, we evaluate Algorithm~\ref{Adaptive_Langive_Samp_algo} within the SGM architectures of~\cite{song2021scorebasedsde,Karras2022ElucidatingTD}. Experiments are conducted on CIFAR-10 ($32\times32$ resolution)~\cite{krizhevsky2009learning_cifar10}, {CelebA~\cite{liu2015deep_celeba}}, FFHQ ($64\times64$)~\cite{Karras2018ASGffhq}, ImageNet ($64\times64$)~\cite{Deng2009ImageNetAL}, and LSUN Bedroom ($256\times256$)~\cite{yu15lsun}. All samplers follow the VE time-discretization strategy~\cite{song2021scorebasedsde} to solve the SDE. For graph generation, we build on SGMs from prior work~\cite{Jo2022ScorebasedGMgraph} and evaluate on benchmark datasets including Ego-small~\cite{Niu2020PermutationIG}, Community-small~\cite{Luo2021GraphDFAD}, and Grid.

\textbf{Baselines.} 
{
For non-SDE frameworks such as VE and VP, we use AMS as the sampler and compare against ALS~\cite{Song2019GenerativeMB}, LDFM~\cite{dockhorn2021critically}, as well as GAN-, probability flow-, and VAE-based methods with pre-trained models~\cite{song2020improved,song2020ddim,ebm}.
}
For NCSN++ (VE-SDE) framework~\cite{song2021scorebasedsde}, we incorporate several SDE solvers, including the Reverse Diffusion (RD) predictor and the Euler–Maruyama (EM) solver~\cite{kloeden1992euler_maru}, both combined with our momentum sampler and the original Langevin Corrector (LC). {
We further compare against DDIM~\cite{song2020ddim}, Heun’s method~\cite{Karras2022ElucidatingTD}, Langevin Dynamics with Fixed Momentum (LDFM), MILD~\cite{shetty2024momentum}, Gotta Go Fast~\cite{jolicoeurmartineau2021gotta}, DPM-Solver-2~\cite{Lu2022DPMSolverAF} on VP model~\cite{song2021scorebasedsde} and report FID values for Analytic-DPM~\cite{bao2022analyticdpm}
}.
For graph generation, we compare against autoregressive models such as GraphRNN~\cite{You2018GraphRNNGR}, DeepGMG~\cite{Li2018deepgmg}, GRAN~\cite{Liao2019EfficientGGran}, and GraphDF~\cite{Luo2021GraphDFAD}, as well as diffusion-based methods including EDP-GNN~\cite{Niu2020PermutationIG}, GDSS~\cite{Jo2022ScorebasedGMgraph}, and DiGress~\cite{vignac2023digress}.

\textbf{Evaluation Metrics.} 
For image generation, we evaluate performance using the Number of Function Evaluations (NFE) and the Fréchet Inception Distance (FID; lower is better, computed with 50,000 samples by default)~\cite{heusel2017gans}.  
{
For graph generation, we evaluate quality using MMD metrics on node degree (Deg.), clustering coefficient (Clus.), and 4-node orbit counts (Orbit), which capture complementary local and higher-order structural properties and are widely used in prior graph generation studies~\cite{Niu2020PermutationIG,Jo2022ScorebasedGMgraph,vignac2023digress} for fair comparison.
}


\subsection{Image Generation Tasks}

{
In this section, we evaluate our Adaptive Momentum Sampler (AMS, Algorithm~\ref{Adaptive_Langive_Samp_algo}) and {Momentum Corrector} (MC, Algorithm~\ref{algo::RDP_MC_VE}) methods in combination with different predictors, including RD and EM. We focus on image synthesis tasks on CIFAR-10, CelebA, FFHQ, and ImageNet with SGMs.
}

\begin{table}[!ht]
    \centering
    \renewcommand{\arraystretch}{1}
    \setlength{\tabcolsep}{10pt}
    \caption{{CIFAR-10 generative performance on VE and VP.}}
    \label{tab:NCSN2_NCSN_IS_FID}
    \begin{tabular}{l l c}
        \toprule
         Model & Sampler & FID $\downarrow$\\
        \midrule
        \multirow{2}{*}{\textbf{VE}} & \textbf{AMS} (ours) &  $9.82$ \\
        &  ALS \cite{Song2019GenerativeMB} & $8.74$ \\
        \midrule
        \multirow{2}{*}{\textbf{VP}} & \textbf{AMS} (ours) & $3.84$ \\
        &  ALS \cite{Song2019GenerativeMB} & $4.24$ \\
        \midrule
       \multirow{2}{*}{\textbf{EBM}} & \textbf{AMS} (ours) & $34.1$ \\
       &  ALS \cite{Song2019GenerativeMB} & $38.7$ \\
        \midrule
        \multirow{5}{*}{\textbf{GANs}} & SNGAN \cite{miyato2018sngan} &$21.7$\\
        & AutoGAN \cite{gong2019autogan}& $12.4$\\
        & DGflow \cite{Ansari2020RefiningDGsngan}&  $9.62$ \\
        & Style \cite{karras2020StyleGAN2}&  $8.32$\\
        & TransGAN \cite{jiang2021transgan}&  $9.26$\\
        \midrule
        \multirow{5}{*}{{\textbf{Flows \& VAEs}}} & Glow \cite{Kingma2018GlowGF} &  $48.9$\\
        &Residual Flow \cite{Chen2019ResidualFlow} &  $46.3$\\
        &NVAE \cite{Vahdat2020NVAEAD} & $23.5$\\
        &DC-VAE \cite{Parmar2020DualCG} & $17.9$\\
        &NCP-VAE \cite{aneja2021ncpvae} & $24.0$\\
        \bottomrule
    \end{tabular}
\end{table}

\paragraph{Experiments on NCSN (VE)} 
We first evaluate our sampler on the NCSN (VE) score network with 500 NFEs. To ensure a fair comparison, we fix the total number of sampling steps and vary $n_\sigma \in \{1, 5\}$. As shown in Table~\ref{tab:ALS_nsigma}, increasing $n_\sigma$ from 1 to 5 consistently reduces FID across all tested methods. Moreover, AMS and ALS exhibit nearly identical sampling times under the same NFE setting, indicating that the momentum term in AMS introduces negligible computational overhead. Under the $n_\sigma = 5$ setting, {
the results in Tables~\ref{tab:NCSN2_NCSN_IS_FID} and~\ref{tab:VE_VP_celeba} show that AMS achieves higher-quality images than all baselines.
}

\begin{table}[!ht]
    \centering
    \renewcommand{\arraystretch}{1}
    \setlength{\tabcolsep}{15pt}
    \caption{{CelebA generative performance on VE and VP models.}}
    \label{tab:VE_VP_celeba}
    \begin{tabular}{l l c}
        \toprule
         Model & Sampler & FID $\downarrow$\\
        \midrule
        \multirow{2}{*}{\textbf{VE}} & \textbf{AMS} (ours) &  $9.30$ \\
        &  ALS \cite{Song2019GenerativeMB} & $10.63$ \\
        \midrule
        \multirow{2}{*}{\textbf{VP}} & \textbf{AMS} (ours) & $6.31$ \\
        &  ALS \cite{Song2019GenerativeMB} & $8.77$ \\
        \bottomrule
    \end{tabular}
\end{table}

\paragraph{Experiments on NCSN++ (VE-SDE)} 
{
We next investigate NCSN++ (VE-SDE) with different predictors. Specifically, we evaluate the FID scores for $n_\sigma \in \{1, 2, 3\}$ using the RD and EM predictors, with the results shown in Figure~\ref{fig:150_NFE_nsigma_snr}. As illustrated in Figure~\ref{fig:150_NFE_nsigma_snr}, the LC sampler yields higher FID scores than the RD sampler under the same NFE settings across all tested step sizes. In contrast, our method consistently outperforms both the EM and RD samplers within an appropriate range of step sizes. Similar to standard neural network training practices, selecting a step size that is too small may lead to slow convergence, whereas an excessively large step size can result in divergence.
}

For samplers without predictors, we set $n_\sigma = 5$ and vary the initial step size $\epsilon$. Our findings reveal that incorporating MC steps consistently improves FID for both EM and RD predictors. In contrast, adding LC correction steps produces mixed results: in small-NFE regimes, LC often degrades FID performance. The best overall performance for EM and RD predictors is achieved with $n_\sigma = 2$. For predictors combined with LC correctors, we follow the recommended choice of $n_\sigma = 1$ and tune $\epsilon$ accordingly.

\begin{table}[!t]
    \centering   
    \setlength{\tabcolsep}{8pt}
    \renewcommand{\arraystretch}{1.2}
    \caption{\textbf{FIDs $\downarrow$ for different NFEs on CIFAR-10, VE-SDE.} Results marked with $^*$ indicate values reported in prior works.} \label{tab::NCSN++_DDPM++_different_NFE}
    \resizebox{\linewidth}{!}{\begin{tabular}{lcccccc}
    \toprule
    Sampler / NFE  & 150  & 195  & 210 & 270 & 300\\
        \hline
      \textbf{RD-MC} (ours) & $6.75 $ & $ \textbf{3.05}$ &$ \textbf{2.86}$ & $\textbf{2.60}$ &$2.61$\\
      RD-LDFM
      & $ 38.6 $ & $ 5.04$& $ 4.37$&$ 4.80$ & $ 4.43$\\
      \hline
       Sampler / NFE &150 & 270 & 500 & 1000 & 2000\\
        \hline
      \textbf{MC} (ours) &$235$ & $27.6$&$9.16$ & $17.7$&$9.29$\\
      RD \cite{song2021scorebasedsde} 
      & $40.3$ & $7.51$& $3.72$& $2.98$& $2.89$\\
      LC \cite{song2021scorebasedsde} 
      & $286$ & $202$& $121$ &$57.8$ & $25.9$\\
      EM \cite{song2021scorebasedsde} 
      & $13.6$ & $12.1$& $12.3$ &$12.6$ &$12.9$\\
      LDFM 
      & $273$& $27.9$&$23.7$ & $32.4$& -\\
      MILD$^*$ \cite{shetty2024momentum} 
      & $20.5$ & $14.5$ &$ 11.5$ & $ 12.1$ & -\\
      RD-LC \cite{song2021scorebasedsde} 
      & $47.6$ &$8.87$ & $3.62$ & $2.62$& $\textbf{2.38}$\\
      Heun's 2nd \cite{Karras2022ElucidatingTD} 
      & $39.6$& $8.63$& $5.01$& $4.16$& $4.06$ \\
      Gotta Go Fast \cite{jolicoeurmartineau2021gotta} 
      & $8.85$ & $3.23$ & $2.87$ & -& -\\
      PIDLD \cite{chen2025pid} & 277 & 33.9 & 11.0 & 9.77 & 9.35\\
      DDIM (VP, ODE) \cite{song2020ddim} 
      & ${4.03}$ & $ 3.53$ & $3.26$ &$3.09$ & $3.01$\\
      Analytic-DPM (VP)$^{*}$ \cite{bao2022analyticdpm} 
      & $5.14$ & $3.55$ &$3.72$ & $4.03$ & - \\
      DPM-Solver (VP, ODE) \cite{Lu2022DPMSolverAF} & $\textbf{3.34}$ & $3.14$ & $2.97$ & $2.93$ & -\\
      \bottomrule
    \end{tabular}}
\end{table}   

Table~\ref{tab::NCSN++_DDPM++_different_NFE} demonstrates that MC acts as an effective corrector for the RD predictor, outperforming RD and EM without correctors. The improvements from MC are also more substantial than those from {LDFM}. For instance, at 150 NFEs, MC reduces the FID for RD from $40.3$ to $6.75$, while LC correction degrades it to $47.6$. {For small NFEs (e.g., 150 and 195), RD-MC achieves FID scores of $6.75$ and $3.05$, respectively.} At 150 NFEs, only DDIM, DPM-solver, and Analytic-DPM outperform RD-MC. This aligns with the design of DDIM, which explicitly targets low-NFE sampling by relaxing the Markov property and accelerating denoising through an ODE-based solver.

We further validate RD-MC on CIFAR-10 using IS and KID under the same settings. As shown in Table~\ref{tab:NCSN++_DDPM++_different_NFE_IS}, MC not only improves FID but also achieves competitive or superior results in IS and KID. This confirms that MC acts as an efficient accelerator while preserving image quality across multiple evaluation metrics.

\begin{table}[!ht]
    \caption{IS $\uparrow$, KID ($\times 10^{-3}$) $\downarrow$ for different NFEs on CIFAR-10, VE-SDE.}\label{tab:NCSN++_DDPM++_different_NFE_IS}
    \centering
    \renewcommand{\arraystretch}{1.2}
    \resizebox{\linewidth}{!}{\begin{tabular}{lcc|cc|cc|cc|cc}
    \toprule
    Sampler/NFE & \multicolumn{2}{c|}{150}  & \multicolumn{2}{c}{195}  & \multicolumn{2}{c}{210} & \multicolumn{2}{c}{270} & \multicolumn{2}{c}{300}\\
    \midrule
     \textbf{RD-MC} (ours) & 9.07 &\textbf{3.13} &  9.51& \textbf{1.06} & 9.55& \textbf{1.04} & \textbf{9.69} & \textbf{0.91} & \textbf{9.70} &\textbf{0.86} \\
        RD-LDFM & 7.69 & 25.3 & 9.25 &2.99 & 9.15&2.27 & 9.32&3.22 &9.25&2.13\\
       \midrule
       Sampler/NFE & \multicolumn{2}{c}{150} & \multicolumn{2}{c}{270} &\multicolumn{2}{c}{500} & \multicolumn{2}{c}{1000} &\multicolumn{2}{c}{2000}\\
        \hline
      \textbf{MC} (ours) & 2.77 &246& 8.65&19.7 &9.97&7.16 & 10.1 &13.2& 9.74 & 5.71\\
      RD & 7.71 & 29.8& 9.17 & 4.11& 9.39 & 1.55& 9.40 & 1.22& 9.46 & 1.25\\
      LC &1.85 & 308 & 3.31 & 198 & 5.37 & 99.6 & 7.55 & 41.1 & 9.20 & 16.9\\
      EM & 9.54 & 11.1 & 9.89 & 10.6& 10.0 & 11.0&10.2 & 11.4 &10.1 & 1.70\\ 
      LDFM & 2.03 & 298& 9.45 & 24.0& 9.63 & 17.4& 9.08 & 18.4 &- & -\\
      RD-LC & 7.46 & 34.9 &9.11 & 4.68& 9.56 & 1.25& 9.73 & 0.60& 9.87 & 0.48\\
      Heun's 2nd & 7.65 & 31.3 & 9.30 & 3.71 & 9.41 & 3.07 & 9.47 & 2.71 & 9.36 & 2.56 \\
      Gotta Go Fast & \textbf{9.87} & -& \textbf{9.64} & - & \textbf{9.57} & -& - & -& - & -\\
      \bottomrule
    \end{tabular}}
\end{table}

\paragraph{Generalization Across Datasets and SGMs}
Finally, we extend our evaluation to FFHQ, ImageNet, and LSUN Bedroom synthesis using the EDM framework~\cite{Karras2022ElucidatingTD, dhariwal2021diffusionmodelsbeatgans}. We compare against RD-LC, DDIM, and Analytic-DPM baselines, with results summarized in Table~\ref{tab:ffhq_imagenet}. Our experiments demonstrate that the proposed method consistently accelerates convergence and improves generation quality, not only on CIFAR-10 but also across diverse datasets and SGMs, extending well beyond NCSN-based models.

\begin{table}[!ht]
    \centering
    \setlength{\tabcolsep}{5pt}
    \renewcommand{\arraystretch}{1}
    \caption{FIDs for different NFEs on FFHQ, ImageNet, and Bedroom.}
    \label{tab:ffhq_imagenet}
    \resizebox{\linewidth}{!}{\begin{tabular}{ccccccc}
    \toprule
       Dataset & NFE  &  \textbf{RD-MC} & RD-LC  & DDIM & Analytic-DPM\\
       \midrule
      \multirow{3}{*}{FFHQ} & 150 & $11.2$& $50.0$ & \textbf{10.9} & 11.6\\
       & 300 & \textbf{9.39} & $13.4$ & 10.1 & 9.88 \\
       & 450 & $\textbf{7.38}$ & $7.50$ & 8.75 & 7.47 \\
       \midrule 
      \multirow{3}{*}{ImageNet} & 150 & \textbf{5.46}& $59.7$& 6.44 & 10.2\\
       & 300 & $\textbf{3.52} $& $11.3$& 5.08 & 7.21 \\
       & 450 & $\textbf{2.90} $& $5.17$& 4.11 & 3.66 \\
        \midrule
       \multirow{2}{*}{Bedroom} & 150 & 4.74 & 32.8 & \textbf{2.46} & -\\
       & 300 & \textbf{2.21} & 15.2 & 2.24 & -\\
       \bottomrule
    \end{tabular}}
\end{table}

\begin{table*}[t]
    \centering
    \renewcommand{\arraystretch}{1.2}
    \small
    \caption{\textbf{Quantitative results $\downarrow$ on generic graph datasets.}}
    \label{tab:graph_overall_tabel}
    \resizebox{.95\linewidth}{!}{\begin{tabular}{ll cccc|cccc|cccc}
    \toprule
         \multirow{2}{*}{\textbf{Dataset}}&  & \multicolumn{4}{c}{\textbf{Ego-small}} & \multicolumn{4}{c}{\textbf{Community-small}} & \multicolumn{4}{c}{\textbf{Grid}}\\
         &  & \multicolumn{4}{c}{Real, $4\leq |V| \leq 18 $} & \multicolumn{4}{c}{Synthetic, $12\leq |V| \leq 20 $} & \multicolumn{4}{c}{Synthetic, $100 \leq |V| \leq 400 $}\\
         \midrule
         & Methods & Deg. $\downarrow$ & Clus. $\downarrow$ & Orbit $\downarrow$& Avg. & Deg. $\downarrow$& Clus. $\downarrow$& Orbit $\downarrow$&  Avg. & Deg. $\downarrow$& Clus. $\downarrow$& Orbit $\downarrow$&  Avg.\\
         \midrule
         \multirow{4}{*}{\textbf{Autoreg.}} & GraphRNN 
         & $0.060$ & $ 0.220$ & $ 0.003$ & $0.104$ & $ 0.080$ & $0.120$ & $0.040$ & $0.080$ & $\textbf{0.064}$ & $0.043$ & $\textbf{0.021}$ & $\textbf{0.043}$\\
         & DeepGMG 
         & $0.040$ & $ 0.100$ & $ 0.020$ & $0.053$ & $ 0.220$ & $0.950$ & $0.400$ & $0.523$ &$ -$ &$ -$ &$ -$ &$ -$\\
         & GRAN 
         & $0.020$ & $ 0.126$ & $ \underline{0.010}$ & $0.052$ & $ 0.070$ & $0.045$ & $0.021$ & $0.045$ &$ -$ &$ -$ &$ -$ &$ -$\\
         & GraphDF 
         & $0.040$ & $ 0.130$ & $\underline{0.010}$ & $0.060$ & $ 0.060$ & $0.120$ & $0.030$ & $0.070$ &$ -$ &$ -$ &$ -$ &$ -$\\
         \midrule
         \multirow{6}{*}{\textbf{Diffusion}} & \textbf{MC} (ours, 300 NFE) & $0.024$& $\textbf{0.009}$& $0.013$&$\textbf{0.015}$ & $0.067$&$0.109$ &$0.031$ & $0.069$ &$ -$ &$ -$ &$ -$ &$ -$\\
         & \textbf{MC} (ours, 750 NFE) & $0.024$& $\underline{0.012}$& $0.014$ & $0.020$ & $\textbf{0.030}$ & $\underline{0.042}$ & $\textbf{0.003}$ & $\textbf{0.025}$ & $ 0.127 $ & $ 0.016 $ &$ 0.203 $ & $ 0.115 $\\
         & EDP-GNN (1000 NFE) 
         & $0.052$ & $ 0.093$ & $ \textbf{0.007}$ & $0.051$ & $ 0.053$ & $0.144$ & $0.026$ & $0.074$& $0.455$ & $0.238$ & $0.328$& $0.340$\\
         &  GDSS (1000+ NFE) 
         & $\textbf{0.021}$ & $ 0.024$ & $ \underline{0.007}$ & $\textbf{0.017}$ & $ 0.045$ & $0.086$ & $0.007$ & $0.046$& $0.378$& $\underline{0.010}$ & $0.442$ &$0.276$\\
         &  GDSS-seq (1000+ NFE) 
         & $0.032$ & $ 0.027$ & $ 0.011$ & $0.023$ & $ 0.090$ & $0.123$ & $0.007$ & $0.073$ & $\underline{0.111}$& $\textbf{0.005}$ & $\underline {0.070}$ &$\underline{0.062}$\\
         & DiGress (500 NFE) 
         &$ 0.120$ & $0.180$ & $0.034$ &$ 0.111$ & $0.089$ & $\textbf{0.019}$ & $0.748$ & $0.285 $& $0.957$& $0.026$ & $1.030$ &$0.671$\\
         \bottomrule
    \end{tabular}}
\end{table*}

\begin{table}[!ht]
    \centering
    \renewcommand{\arraystretch}{1.1}
    \setlength{\tabcolsep}{5.8pt}
    \caption{{Quantitative results on Community-small.}}
    \label{tab:graph_NFE_comp}
    \begin{tabular}{clccccc}
    \toprule
        NFE & Methods & Deg. & Clus. & Orbit &  Time/s\\
        \midrule
        \multirow{4}{*}{300} & EM & $0.074$& $0.111$& $\underline{0.007}$& $2.99$&\\
        & EM-LD& $ \underline{0.061}$& $ 0.122$& $ 0.031$ & $ 2.86$\\
        & EM-LDFM & $0.065$& $0.122$&$0.031$ &$2.94$ \\
         & EM-MC (ours)& $0.067$&$\underline{0.109}$ &$0.031$ & $2.83$ \\
         \midrule
         \multirow{4}{*}{450} & EM & $0.049$& $0.121$& $0.013$& $4.28$\\
         & EM-LD & $0.042$& $0.071$& $0.005$& $4.09$\\
         & EM-LDFM & $0.043$& $0.068$&$0.005$ &$4.27$ \\
         & EM-MC (ours)&  $\underline{0.041}$& $\underline{0.053}$& $\underline{0.004}$& $4.10$\\
         \midrule
         \multirow{4}{*}{600} & EM & $0.066$& $0.173$& $0.011$& $5.19$\\
         & EM-LD & $0.025$& $0.124$&$\underline{0.004}$ & $5.21$\\
         & EM-LDFM & $0.046$& $0.092$&$0.006$ &$5.40$ \\
         & EM-MC (ours)& $\underline{0.037}$& $\underline{0.048}$& $\underline{0.004}$& $5.26$\\ 
         \midrule
         \multirow{4}{*}{750} & EM & $0.451$& $0.123$& $0.006$& $6.63$\\
         & EM-LD & $0.050$& $0.065$&$0.004$ &$6.41$ \\
         & EM-LDFM& $0.047$& $0.072$&$\underline{\textbf{0.003}}$ &$7.03$ \\
         & EM-MC (ours)&$\underline{\textbf{0.030}}$ & $\underline{\textbf{0.042}}$ & $\underline{\textbf{0.003}}$ & $6.88$\\
    \bottomrule
    \end{tabular}
\end{table}

\subsection{Graph Generation Tasks}

We extend Algorithm~\ref{Adaptive_Langive_Samp_algo} to graph generation in order to further validate the performance and generality of our method. For hyperparameter tuning, we search the initial sampling step size $\epsilon \in \{0.5, 0.6, 0.7, 0.8, 0.9, 1.0\}$ and select the value yielding the lowest average metric. Following our previous analysis, we set $n_{\sigma} = 2$ as the default.

\paragraph{Overall Performance}
Table~\ref{tab:graph_overall_tabel} reports the results on several benchmark datasets with NFEs of 300 and 750. On the \textbf{Ego-small} dataset, 300 NFEs are sufficient for our MC sampler to achieve competitive performance, yielding a speedup of more than $3\times$ compared to baseline diffusion samplers. On the larger \textbf{Community-small} dataset, 750 NFEs suffice for our method to generate high-quality samples, corresponding to a speedup of approximately $2\times$.

We also evaluate on the \textbf{Grid} dataset ($100 \le |V| \le 400$). Here, only GDSS-seq~\cite{Jo2022ScorebasedGMgraph} and GraphRNN~\cite{You2018GraphRNNGR} achieve better quantitative results. Nevertheless, with 750 NFEs, our method achieves performance comparable to GDSS-seq with about $2\times$ faster sampling. It is worth noting that GDSS-seq generates node features and the weighted adjacency matrix sequentially, effectively doubling its computational cost. Taken together, these results demonstrate that our MC sampler provides a general and efficient solution for graph generation.

\paragraph{Comparison with Other Correctors}
Beyond direct comparisons with existing methods, we further evaluate our sampler against alternative SDE correctors, including LD, LDFM, and the no-corrector setting, across different NFE budgets. We adopt the same metrics and measure the time required to generate 128 graphs on a single NVIDIA 4090 GPU. The results are summarized in Table~\ref{tab:graph_NFE_comp}. We observe that the additional $\beta$ computation in our method has a negligible impact on runtime. As NFE increases, our MC consistently outperforms all tested Markov samplers across all metrics, while maintaining sampling times close to those of the EM-LD baseline.

{
\subsection{Additional Investigation}
\paragraph{Study the effect of $\delta$.} Our adaptive momentum mechanism introduces a hyperparameter, $\delta$. To evaluate the robustness of our method with respect to this parameter, we conduct an ablation study. Following the experimental setup described in Section~\ref{sec::numerical results}, we use the VE-SDE sampler to generate images on the CIFAR-10 and ImageNet datasets with 150 and 300 NFEs, and evaluate the results using the FID score. Specifically, we consider $\delta \in \{10^{-1}, 10^{-2}, 10^{-3}, 10^{-4}, 10^{-5}\}$. The experimental results are reported in Table~\ref{tab:ablation_delta}.
\begin{table}[!ht]
    \centering
    \setlength{\tabcolsep}{5pt}
    \renewcommand{\arraystretch}{1}
    \caption{{
    Ablation study on hyperparameter $\delta$.
    }}
    \label{tab:ablation_delta}
    \resizebox{\linewidth}{!}{\begin{tabular}{c|ccccccc}
    \toprule
       Dataset & NFE  &  $10^{-1}$ & $10^{-2}$  & $10^{-3}$ & $10^{-4}$ & $10^{-5}$\\
       \midrule
      \multirow{2}{*}{CIFAR-10} & 150 & $ 7.21$& $6.76$ & $6.75$ & $6.75$ & $6.75$\\
       & 300 & $3.14$ & $2.66$ & $2.61$ & $2.61$ & $2.61$\\
       \midrule 
      \multirow{2}{*}{ImageNet} & 150 & $5.51 $&$5.46$& $5.46$& $5.46 $ & $5.46$\\
       & 300 &$4.18 $ &  $3.66$ & $3.52$& $3.52$ & $3.52$ \\
       \bottomrule
    \end{tabular}}
\end{table}}
{
As shown in Table~\ref{tab:ablation_delta}, our algorithm is largely insensitive to the choice of $\delta$. In particular, when $\delta \leq 10^{-3}$, the resulting FID scores remain identical. This observation indicates that the adaptively computed momentum $\beta_t$ remains below approximately $0.999$ in most cases. Performance degradation is observed only when the momentum upper bound is restricted to $0.9$ (i.e., $\delta = 0.1$). In all experiments reported in the main text, we therefore set $\delta = 10^{-3}$ by default, further demonstrating the hyperparameter-free property of our adaptive momentum mechanism.
}

{
\paragraph{Computational Complexity.}
    We further analyze the per-step computational complexity of the proposed adaptive momentum sampling method. Similar to standard Langevin sampling, the dominant computational cost arises from the forward evaluation of the score network $s_\theta(\mathbf{x},\sigma)$, which requires $\mathcal{O}(C_{\text{net}})$ floating-point operations (FLOPs), where $C_{\text{net}}$ denotes the complexity of the neural network.}
    {
    Compared with the standard Langevin update, our method introduces additional operations for computing the adaptive momentum parameter $\beta$. Specifically, computing $\beta$ requires evaluating two vector differences $(\mathbf{g}^t-\mathbf{g}^{t-1})$ and $(\mathbf{x}^t-\mathbf{x}^{t-1})$, which costs $2d$ FLOPs, and two $\ell_2$ norms, each requiring approximately $2d$ FLOPs (including elementwise squares and summation). Therefore, the computation of $\beta$ requires approximately $6d$ FLOPs. The remaining operations, including scalar arithmetic and projection, contribute only a negligible constant cost.}

    {
    As a result, the per-step computational complexity of our method can be written as
    \[
    C_{\text{AMS}} = C_{\text{net}} + 6d + \mathcal{O}(1),
    \]
    where $d$ denotes the dimensionality of the data. In comparison, the standard Langevin update requires approximately
    \[
    C_{\text{Langevin}} = C_{\text{net}} + \mathcal{O}(d).
    \]
    In practical diffusion and score-based generative models, the neural network evaluation dominates the total cost (i.e., $C_{\text{net}} \gg d$). Consequently, the additional $6d$ operations introduced by the adaptive momentum computation represent a negligible overhead. This theoretical analysis is consistent with the empirical runtime results reported in Tables~\ref{tab:ALS_nsigma} and~\ref{tab:graph_NFE_comp}.
}

\section{Conclusion}
Motivated by traditional SGD optimization, we propose a novel gradient-based sampling method leveraging the Markov technique for existing SGMs. This sampler requires no additional information beyond the score function, making it broadly applicable. Theoretically, we analyze the convergence properties and efficiency within the convex optimization framework using Markov Chains, providing an alternative perspective to the conventional SDE/ODE-based analysis in SGM sampling. Our method enables the examination of data distribution properties during sampling, such as computing the mean and variance at each step, revealing the sampling distribution's evolution with the sampling time step $t$. By bridging SGD and gradient-based sampling, our approach opens new avenues for accelerating SGM applications. Experimentally, we validate our method across various image and graph generation tasks using different datasets and SGMs. Our results demonstrate that it significantly enhances SGM performance in reverse SDE sampling without incurring substantial additional overhead. This underscores its broad applicability across diverse tasks and models. Overall, our Markov Chain-based sampler presents distinct advantages for SGM sampling, offering both efficiency and scalability.

\section{Discussion}
{We outline several potential directions for our work here.
\paragraph{Potential Extension to ODE-based Samplers}
Although our method is developed for stochastic samplers, the adaptive momentum mechanism is broadly applicable and may also be extended to deterministic ODE-based samplers, such as flow-matching generative models. Existing momentum-accelerated ODE methods (e.g.,~\cite{Wang2023BoostingDM}) improve convergence but need additional hyperparameters, making practical tuning more difficult. In contrast, our adaptive momentum approach introduces no additional hyperparameters while still providing acceleration. This suggests that similar adaptive strategies could be incorporated into ODE-based samplers to improve efficiency. Investigating this extension could help improve the efficiency and stability of ODE samplers in future studies.
\paragraph{Broader Evaluation}
In this work, we primarily focus on image and graph generation, demonstrating the effectiveness of our adaptive momentum method across a range of models and datasets. SGMs have also been successfully applied to text generation \cite{Li2022DiffusionLMIC_text} and video generation \cite{Ho_vedio_diffusion}. Extending our adaptive momentum sampling approach to these tasks presents both challenges and opportunities. On the one hand, conditioning on high-dimensional text embeddings may introduce additional complexity, affecting stability and convergence. On the other hand, our momentum-based approach is plug-and-play and requires no additional hyperparameters, which could help accelerate conditional sampling while maintaining high generation quality~\cite{sun2026convergence,sun2026gradient}.
}

\section*{Acknowledgements}
The authors sincerely thank the Editor and the anonymous reviewers for their careful evaluation of our manuscript and for their constructive suggestions.

\bibliography{reference}
\bibliographystyle{IEEEtran}

\appendices
\setcounter{page}{1}
\setcounter{proposition}{0}

\section{Lemmas and Proofs}\label{app:lemma}

\begin{lemma} \cite{Dieuleveut2017BridgingTG} \label{lemma_int=0}
     We have that for any $\alpha \in (0,1/L)$, the integral $\int_{\mathbb R^d}\nabla f(\mathbf{x})\pi_{\alpha}(d\mathbf{x}) = 0$.
\end{lemma}

\begin{lemma}[\textbf{Strong convexity case}] \cite{wennormalized} \label{lemma_strconvex}
\textit{Let $f(\mathbf{x})$ be a strongly convex function and {twice differentiable}, and let ${\{\mathbf{x}^k}\}_{k\ge 0}$ be generated by Eq. \eqref{eq::NSHBupd}. Assume that $$0<\mu\le\min_{\mathbf{x}}\{\lambda_{\min}(\nabla^2 f(\mathbf{x}))\}\le \max_{\mathbf{x}}\{\lambda_{\max}(\nabla^2 f(\mathbf{x}))\}\le L.$$ Then we have}
$$
\mathbb{E}\|{\mathbf{x}^K}-{\mathbf{x}^*}\|^2= (1-2\alpha\delta\mu)^K\mathbb{E} \|{\mathbf{x}^1-\mathbf{x}^*}\|^2 + \mathcal{O}(\alpha).
$$
\end{lemma}

This Lemma indicates that $\int_{\mathbb R^d}\|\mathbf{x}-\mathbf{x}^{*}\|^2\pi_{\alpha}d(\mathbf{x}) = \mathcal{O}({\alpha})$ if $\mathbf{x}^{0}$ is distributed according to $\pi_{\alpha}^*$ and the Markov Kernel is stationary. Also, we have
$\int_{\mathbb R^d}\|\mathbf{x}-\mathbf{x}^{*}\|^3\pi_{\alpha}d(\mathbf{x}) = \mathcal{O}(\alpha^{3/2})$. 

\begin{lemma}\label{lemma:best_beta_lemma}\cite{wennormalized}
    Let $f$ be a quadratic function with the form {$f({\mathbf{x}})=\frac{1}{2}{\mathbf{x}}^T{\mathbf{A}}{\mathbf{x}}+{\mathbf{b}}^T{\mathbf{x}}+{c}$} and $\nabla f(\mathbf{x})=\mathbf{A}\mathbf{x}+\mathbf{b}$. Moreover, we denote the smallest and the largest eigenvalue of $\mathbf{A}$ as $\mu$ and $L$. Then, with any given fixed $0 <\alpha \le \frac{1}{L}$, the optimal choice for $\beta$ in the normalized SHB method is $\beta = (\frac{1-\alpha \mu}{1+\alpha \mu})^2$. Moreover, normalized SHB can lead to a convergence rate of $\|{\mathbf{x}^k-\mathbf{x}^*}\|\le (\frac{1-\alpha \mu}{1+\alpha \mu}+\epsilon)^k\|{\mathbf{x}^0}-{\mathbf{x}^*}\|$, where $k\ge K({\mathbf{A}},\epsilon)$ and $\epsilon$ is a small enough real number.
\end{lemma}

\begin{lemma}\cite{wennormalized, Sun2021TrainingDN}\label{lemma:approximate_mu}
    Assume the conditions in Lemma \ref{lemma:best_beta_lemma} hold, and let {$\{\mathbf{x}^k\}_{k\ge 0}$} be generated by  \eqref{eq::NSHBupd}. Assume that $\mathbf{A}$ has a unique minimum eigenvalue $\mu$. Then, for any fixed learning rate $\alpha\le\frac{1}{L}$, we can approximate $\mu$ by
    \begin{equation}
        \underset{k\rightarrow \infty}{\lim}\frac{\|{\mathbf{g}}^k-{\mathbf{g}}^{k-1}\|}{\|{\mathbf{x}}^{k}-{\mathbf{x}}^{k-1}\|}=\mu.
    \nonumber
    \end{equation}
\end{lemma}

\subsection{Proof of Proposition \ref{proposition_pi_stationary}}

\begin{proof}
For two distributions {
$\theta,\hat{\theta}\in\mathcal{P}_2(\mathbb{R}^{2d})$
}, there exist random variables $\mathbf{z}^{0}, \hat{\mathbf{z}}^{0}$ such that
\begin{equation}\nonumber
    \mathcal{W}_{2}^{2}(\theta,\hat{\theta}) = \mathbf{E}\left[\|\mathbf{z}^{0}-\hat{\mathbf{z}}^0\|^2\right]
\end{equation}
By Theorem 4.1 of \cite{Villani2008OptimalTO}, we use the synchronous coupling: the two chains initialized at $\mathbf{z}^{0}$ and $\hat{\mathbf{z}}^{0}$ are driven by the same additive noise sequence $\{\boldsymbol{\varepsilon}^{t}\}_{t\ge 1}$. Then the distribution of $(\mathbf{z}^{t},\hat{\mathbf{z}}^t)$ belongs to $\Pi (\theta R_{\alpha}^{t},\hat{\theta}R_{\alpha}^{t})$. Let
\begin{equation}\nonumber
    \mathbf{c}:=
    \begin{bmatrix}
    -\alpha(1-\beta)\mathbf{b} \\
    \mathbf{0}
    \end{bmatrix}.
\end{equation}
Under the synchronous coupling, we derive

{
\begin{equation}\nonumber
\begin{aligned}
 &\mathcal{W}^{2}_{2}(\theta R_{\alpha}^{t},\hat{\theta}R_{\alpha}^{t}) \leq \mathbf{E}\|\mathbf{z}^t- \hat{\mathbf{z}}^t\|^2  
\\ &= \mathbb{E} \|\mathbf{T}\mathbf{z}^{t-1}+\mathbf{c}+\boldsymbol{\varepsilon}^{t}-\mathbf{T}\hat{\mathbf{z}}^{t-1}-\mathbf{c}-\boldsymbol{\varepsilon}^{t}\|^2   \\
&=\mathbb{E}\|\mathbf{T}(\mathbf{z}^{t-1}-\hat{\mathbf{z}}^{t-1})\|^2 =\dots =\mathbb{E}\|\mathbf{T}^t(\mathbf{z}^0-\hat{\mathbf{z}}^0)\|^2 \\ & \leq\|\mathbf{T}^t\|^2\mathbb{E}\|\mathbf{z}^0 - \hat{\mathbf{z}}^0\|^2 =\|\mathbf{T}^t\|^2\mathcal{W}_{2}^2(\theta,\hat{\theta}).
\end{aligned}
\end{equation}
}

{
Using Gelfand's Theorem~\cite{Godsil2001}, for any \(0<\epsilon<1-\rho(\mathbf{T})\), there exists \(C_\epsilon>0\) such that
\[
\|\mathbf{T}^{t}\| \leq C_\epsilon(\rho(\mathbf{T})+\epsilon)^{t}.
\]
Hence,
\[
\mathcal{W}_{2}(\theta R_{\alpha}^{t},\hat{\theta}R_{\alpha}^{t})
\leq
C_\epsilon(\rho(\mathbf{T})+\epsilon)^t
\mathcal{W}_{2}(\theta,\hat{\theta}).
\]
}
This indicates that, for any $\theta, \hat{\theta}$, 
{\begin{equation}\label{different_mu_equal_0}
    \underset{t\rightarrow +\infty}{\lim}\mathcal{W}_2(\theta R_{\alpha}^t, \hat{\theta}R_{\alpha}^t) = 0.
\end{equation}
Let \(r=\rho(\mathbf{T})+\epsilon<1\). For any \(m>n\), by the triangle inequality and the contraction estimate above, we have
\[
\begin{aligned}
\mathcal{W}_{2}(\theta R_{\alpha}^{m},\theta R_{\alpha}^{n})
&\leq
\sum_{k=n}^{m-1}
\mathcal{W}_{2}(\theta R_{\alpha}^{k+1},\theta R_{\alpha}^{k})\\
&=
\sum_{k=n}^{m-1}
\mathcal{W}_{2}((\theta R_{\alpha})R_{\alpha}^{k},\theta R_{\alpha}^{k})\\
&\leq
C_{\epsilon}\mathcal{W}_{2}(\theta R_{\alpha},\theta)
\sum_{k=n}^{m-1}r^{k}\\
&\leq
\frac{C_{\epsilon}r^{n}}{1-r}
\mathcal{W}_{2}(\theta R_{\alpha},\theta).
\end{aligned}
\]
Thus \((\theta R_{\alpha}^{t})_{t\geq0}\) is a Cauchy sequence in
\((\mathcal{P}_{2}(\mathbb{R}^{2d}),\mathcal{W}_{2})\). Since this space is complete, there exists
\(\pi_{\alpha}^{\theta}\in\mathcal{P}_{2}(\mathbb{R}^{2d})\) such that
\[
\lim_{t\to+\infty}\mathcal{W}_{2}(\theta R_{\alpha}^{t},\pi_{\alpha}^{\theta})=0.
\]
} In the following, we prove $\pi_{\alpha}^{\theta}$ does not depend on $\theta$ and is unique. Assume that there exists a distribution $\pi_{\alpha}^{\hat{\theta}}$ that satisfies $\underset{t\rightarrow +\infty}{\lim} \mathcal{W}_{2}(\hat{\theta}R_{\alpha}^{t},\pi_{\alpha}^{\hat{\theta}}) = 0$.
By the triangle inequality, we have
\begin{equation}\nonumber
    \mathcal{W}_{2}(\pi_{\alpha}^{\theta},\pi_{\alpha}^{\hat{\theta}})\leq \mathcal{W}_{2}(\pi_{\alpha}^{\theta},\theta R_{\alpha}^{t})+\mathcal{W}_{2}(\theta R_{\alpha}^{t},\hat{\theta}R_{\alpha}^{t})+\mathcal{W}_{2}(\hat{\theta}R_{\alpha}^{t},\pi_{\alpha}^{\hat{\theta}}).
\end{equation}
Taking the limits as \(t\rightarrow +\infty\) and using Eq.~\eqref{different_mu_equal_0}, we conclude $ \mathcal{W}_{2}(\pi_{\alpha}^{\theta},\pi_{\alpha}^{\hat{\theta}}) = 0$. We see that $\pi_{\alpha}^{\theta} = \pi_{\alpha}^{\hat{\theta}}$. Therefore, we can denote $\pi_{\alpha}^{\theta}$  as $\pi_{\alpha}^{*}$. We can see that $\pi_{\alpha}^{*}$ is unique for given $\alpha$.
{
Moreover, since \(R_{\alpha}\) is induced by the affine map with additive noise in Eq.~\eqref{markovz}, it is continuous with respect to \(\mathcal{W}_{2}\). Hence, 
\[
\theta R_\alpha^{t+1}=(\theta R_\alpha^t)R_\alpha \to \pi_\alpha^*R_\alpha.
\]
On the other hand, \(\theta R_\alpha^{t+1}\to \pi_\alpha^*\). By uniqueness of the limit in \(\mathcal W_2\), we obtain
\[
\pi_\alpha^*R_\alpha=\pi_\alpha^*.
\]
} To this end, we begin to investigate the spectral radius of matrix $\mathbf{T}$, denote the eigenvalues of $\mathbf{T}$  by $\lambda$,
\begin{equation}\nonumber
\begin{aligned}
    & \text{det} \begin{bmatrix}
        (1+\beta)\mathbf{I}-\alpha(1-\beta)\mathbf{A}-\lambda\mathbf{I} & -\beta\mathbf{I} \\
        \mathbf{I} & -\lambda\mathbf{I}
    \end{bmatrix} = 0 \\ &\Rightarrow \text{det}\begin{bmatrix}
        (1+\beta-\lambda-\frac{\beta}{\lambda})\mathbf{I}-\alpha(1-\beta)\mathbf{A} & \mathbf{0} \\
        \mathbf{I} & -\lambda\mathbf{I}
        \end{bmatrix}=0 \\
        & \Rightarrow \text{det}
            ((\lambda+\frac{\beta}{\lambda})\mathbf{I}-((1+\beta)\mathbf{I}-\alpha(1-\beta)\mathbf{A}))=0.
\end{aligned}
\end{equation}
Since \(\beta>0\), \(\lambda=0\) is not an eigenvalue of \(\mathbf{T}\). Thus, for any eigenvalue \(\lambda\) of \(\mathbf{T}\), we have \(\lambda\neq0\). This indicates that \(\lambda+\frac{\beta}{\lambda}\) is an eigenvalue of \((1+\beta)\mathbf{I}-\alpha(1-\beta)\mathbf{A}\). We denote by \(\lambda^*\) any eigenvalue of \((1+\beta)\mathbf{I}-\alpha(1-\beta)\mathbf{A}\).
By \cite{wennormalized}, under the prescribed choice of $\beta$, the spectral radius of $\mathbf{T}$ satisfies
\begin{equation}\label{lambda_max_T}
    \rho(\mathbf{T})\leq \frac{1-\alpha\mu}{1+\alpha\mu}<1.
\end{equation}
Putting this together with the preceding contraction argument, we then derive the proposition.

\end{proof}

\subsection{Proof of Proposition \ref{proposition_expec_quard}}
\begin{proof}
We denote {the optimal choice} of $\mathbf{x}$ as $\mathbf{x}^*$, $\hat{\mathbf{z}}^{k+1}$ as $\begin{bmatrix}
    \mathbf{x}^{k+1}-\mathbf{x}^* \\
    \mathbf{x}^{k} -\mathbf{x}^*
\end{bmatrix}$, and $\mathbf{z}^*$ as $\begin{bmatrix}
    \mathbf{x}^*\\
    \mathbf{x}^*
\end{bmatrix}$. Let
\begin{equation}\nonumber
    \mathbf{S}_{\mathbf{z}}:= \underset{\mathbb{R}^{2d}}{\int}(\mathbf{z}-\mathbf{z}^{*})^{\otimes 2}\pi_{\alpha}^{*}(d\mathbf{z}).
\end{equation}
{
Since \(\pi_{\alpha}^{*}\) is stationary for Eq.~\eqref{markovz}, and
\(\boldsymbol{\varepsilon}_{k+1}\) is centered and independent of \(\mathbf{z}^{k}\), we have
\[
    \hat{\mathbf{z}}^{k+1}
    =
    \mathbf{T}\hat{\mathbf{z}}^{k}
    +
    \boldsymbol{\varepsilon}_{k+1}.
\]
}
Let \(\boldsymbol{\Sigma}_{\boldsymbol{\varepsilon}}:=\mathbb E[\boldsymbol{\varepsilon}_{k+1}\boldsymbol{\varepsilon}_{k+1}^{T}]\). Taking the second moment under the stationary distribution yields the discrete Lyapunov equation
\begin{equation}\nonumber
    \mathbf{S}_{\mathbf{z}}=\mathbf{T}\mathbf{S}_{\mathbf{z}}\mathbf{T}^{T}+\boldsymbol{\Sigma}_{\boldsymbol{\varepsilon}}.
\end{equation}
Vectorizing both sides gives
\begin{equation}\nonumber
    (\mathbf{I}-\mathbf{T}\otimes\mathbf{T})\operatorname{vec}(\mathbf{S}_{\mathbf{z}})=
        \operatorname{vec}(\boldsymbol{\Sigma}_{\boldsymbol{\varepsilon}}).
\end{equation}
Since \(\rho(\mathbf{T})<1\), we have \(\rho(\mathbf{T}\otimes\mathbf{T})=\rho(\mathbf{T})^2<1\), hence \(\mathbf{I}-\mathbf{T}\otimes\mathbf{T}\) is invertible. We obtain
\begin{equation}\nonumber
    \operatorname{vec}(\mathbf{S}_{\mathbf{z}})=
    (\mathbf{I}-\mathbf{T}\otimes\mathbf{T})^{-1}\operatorname{vec}(\boldsymbol{\Sigma}_{\boldsymbol{\varepsilon}}).
\end{equation}
\end{proof}

\subsection{Proof of Proposition \ref{proposition_conv}}

\begin{proof}
Let {$\alpha \in (0,1/L)$}, $(\mathbf{x}^k)_{k\ge 0}$ be generated by Algorithm \ref{Adaptive_Langive_Samp_algo}, $\mathbf{x}^{0}$ distributed according to $\pi_{\alpha}$, and \(\mathbf{H}:=\nabla^2 f(\mathbf{x}^*)\). The local linearized transition matrix is \(\mathbf{T}_{\mathbf{H}}\). For given function $f$, we perform a Taylor expansion {around $\mathbf{x}^*$}, then we derive
\begin{equation}\label{eq:taylor_3rd_expand}
    \nabla f(\mathbf{x}) = \mathbf{H}(\mathbf{x}-\mathbf{x}^{*})+\frac{1}{2}\nabla^3 f(\mathbf{x}^{*})(\mathbf{x}-\mathbf{x}^*)^{\otimes 2}+\mathcal{R}_3(\mathbf{x}),
\end{equation}
where $\mathcal{R}_3: \mathbb R^d\rightarrow \mathbb R^d$ satisfies
\begin{equation}\label{3rd_remain}
    \sup_{\mathbf{x}\in \mathbb R^d,\ \mathbf{x}\neq \mathbf{x}^{*}}
    \frac{\|\mathcal{R}_3(\mathbf{x})\|}
    {\|\mathbf{x}-\mathbf{x}^{*}\|^3}
    <+\infty.
\end{equation}
{Using} Lemma \ref{lemma_int=0}, we take the integral on both sides of Eq. \eqref{eq:taylor_3rd_expand}, with respect to $\pi_{\alpha}$, we are led to
{
\small
\begin{equation}\nonumber
    0 = \underset{\mathbb R^d}{\int}\Big(\nabla^2 f(\mathbf{x}^*)(\mathbf{x}-\mathbf{x}^*) +\frac{1}{2}\nabla^3f(\mathbf{x}^*)(\mathbf{x}-\mathbf{x}^{*})^{\otimes 2} +\mathcal{R}_3(\mathbf{x})\Big)\pi_{\alpha}(d\mathbf{x}).
\end{equation}
}
Then, using Lemma \ref{lemma_strconvex} and H\"older's inequality, we are led to
{
\small
\begin{equation}\label{eq:expo_alpha_3_half}
    \mathbf{H}(\overline{\mathbf{x}}-\mathbf{x}^{*})+\frac{1}{2}\nabla^3 f(\mathbf{x}^{*})\Big[\underset{\mathbb R^d}{\int}(\mathbf{x}-\mathbf{x}^{*})^{\otimes 2}\pi_{\alpha}(d\mathbf{x})\Big]=\mathcal{O}(\alpha^{\frac{3}{2}}).
\end{equation}
}
Besides, we use the augmented-state covariance to keep the momentum contribution in the second moment. Let
\begin{equation}\nonumber
 \begin{aligned}
     &\mathbf{P}:=\begin{bmatrix}\mathbf{I}&\mathbf{0}\end{bmatrix},
    \mathbf{S}_{\mathbf{z}}:= \underset{\mathbb R^{2d}}{\int}(\mathbf{z}-\mathbf{z}^{*})^{\otimes 2}\pi_{\alpha}^{*}(d\mathbf{z}),
    \\& \mathbf{S}_{\mathbf{x}}:= \underset{\mathbb R^{d}}{\int}(\mathbf{x}-\mathbf{x}^{*})^{\otimes 2}\pi_{\alpha}(d\mathbf{x}).
 \end{aligned}
\end{equation}
Since $\mathbf{x}-\mathbf{x}^{*}=\mathbf{P}(\mathbf{z}-\mathbf{z}^{*})$, we have
\begin{equation}\nonumber
    \mathbf{S}_{\mathbf{x}}=\mathbf{P}\mathbf{S}_{\mathbf{z}}\mathbf{P}^{T},
    \operatorname{vec}(\mathbf{S}_{\mathbf{x}})=(\mathbf{P}\otimes\mathbf{P})\operatorname{vec}(\mathbf{S}_{\mathbf{z}}).
\end{equation}
Under the local Taylor expansion around $\mathbf{x}^{*}$, the augmented centered recursion can be written as
\begin{equation}\nonumber
    \hat{\mathbf{z}}^{k+1}=\mathbf{T}_{\mathbf{H}}\hat{\mathbf{z}}^{k}+\boldsymbol{\varepsilon}_{\mathbf{z},k+1}+\mathbf{r}_{k+1},
\end{equation}
where $\hat{\mathbf{z}}^{k}:=\mathbf{z}^{k}-\mathbf{z}^{*}$, $\boldsymbol{\varepsilon}_{\mathbf{z},k+1}=\begin{bmatrix}\boldsymbol{\varepsilon}_{\mathbf{x},k+1}\\ \mathbf{0}\end{bmatrix}$, and the remainder satisfies
\begin{equation}\nonumber
\mathbb{E}\|\mathbf{r}_{k+1}\|^{2}
+
\mathbb{E}\left|\left\langle \mathbf{T}_{\mathbf{H}}\hat{\mathbf{z}}^{k}
+\boldsymbol{\varepsilon}_{\mathbf{z},k+1},\mathbf{r}_{k+1}\right\rangle\right|
=
\mathcal{O}(\alpha^{\frac{5}{2}}).
\end{equation}
This follows from Lemma~\ref{lemma_strconvex}, Eq.~\eqref{3rd_remain}, and H\"older's inequality. Taking the second moment under the stationary distribution yields
\begin{equation}\nonumber
    \mathbf{S}_{\mathbf{z}}=\mathbf{T}_{\mathbf{H}}\mathbf{S}_{\mathbf{z}}\mathbf{T}_{\mathbf{H}}^{T}+\boldsymbol{\Sigma}_{\boldsymbol{\varepsilon}}+\mathcal{O}(\alpha^{\frac{5}{2}}),
\end{equation}
where
\begin{equation}\nonumber
    \boldsymbol{\Sigma}_{\boldsymbol{\varepsilon}}:=\mathbb{E}[\boldsymbol{\varepsilon}_{\mathbf{z},k+1}\boldsymbol{\varepsilon}_{\mathbf{z},k+1}^{T}]
    =
    \begin{bmatrix}
    \alpha^2\mathcal{C}(\mathbf{x}^{*})+\mathcal{O}(\alpha^{\frac{5}{2}})&\mathbf{0}\\
    \mathbf{0}&\mathbf{0}
    \end{bmatrix}.
\end{equation}
Equivalently,
\begin{equation}\nonumber
    (\mathbf{I}-\mathbf{T}_{\mathbf{H}}\otimes\mathbf{T}_{\mathbf{H}})\operatorname{vec}(\mathbf{S}_{\mathbf{z}})
    =\operatorname{vec}(\boldsymbol{\Sigma}_{\boldsymbol{\varepsilon}})+\mathcal{O}(\alpha^{\frac{5}{2}}).
\end{equation}
{
Since \(\rho(\mathbf{T}_{\mathbf{H}})<1\), \(\mathbf{I}-\mathbf{T}_{\mathbf{H}}\otimes\mathbf{T}_{\mathbf{H}}\) is invertible. Moreover, in the small-step regime, we use the local expansion
\[
\mathbf{I}-\mathbf{T}_{\mathbf{H}}\otimes\mathbf{T}_{\mathbf{H}}
=
\alpha \mathbf{G} + \mathcal{O}(\alpha^2),
\]
where \(\mathbf{G}\) denotes the leading-order matrix of the local expansion. Under the local nondegeneracy condition that \(\mathbf{G}\) is nonsingular, the \(O(\alpha^{5/2})\) remainder in the vectorized Lyapunov equation becomes \(O(\alpha^{3/2})\) after applying \((\mathbf{I}-\mathbf{T}_{\mathbf{H}}\otimes\mathbf{T}_{\mathbf{H}})^{-1}\).
}

\begin{equation}\nonumber
\begin{aligned}
    \operatorname{vec}(\mathbf{S}_{\mathbf{x}})
    &=(\mathbf{P}\otimes\mathbf{P})(\mathbf{I}-\mathbf{T}_{\mathbf{H}}\otimes\mathbf{T}_{\mathbf{H}})^{-1}
    \operatorname{vec}(\boldsymbol{\Sigma}_{\boldsymbol{\varepsilon}})
    +\mathcal{O}(\alpha^{\frac{3}{2}})\\
    &=(\mathbf{P}\otimes\mathbf{P})(\mathbf{I}-\mathbf{T}_{\mathbf{H}}\otimes\mathbf{T}_{\mathbf{H}})^{-1}
    \operatorname{vec}\left(
    \begin{bmatrix}
    \alpha^2\mathcal{C}(\mathbf{x}^{*})&\mathbf{0}\\
    \mathbf{0}&\mathbf{0}
    \end{bmatrix}
    \right)
    \\&+\mathcal{O}(\alpha^{\frac{3}{2}}).
\end{aligned}
\end{equation}
Combining Eq. \eqref{eq:expo_alpha_3_half} and the above gives Proposition \ref{proposition_conv}.
\end{proof}

\input{samples}

\vfill

\end{document}

%% file: samples.tex
\clearpage
\onecolumn

\subsection{Hyperparameters}\label{app:parameters}

We summarize the hyperparameter settings used in our experiments to facilitate reproducibility.  

\paragraph{NCSN (VE) Sampling.} 

Our AMS sampler was consistently evaluated with $n_\sigma = 5$. We found that using $n_\sigma = 5$ (instead of $n_\sigma = 1$) leads to better FID performance for both ALS and AMS, though the total number of sampling steps must be divided by five when $n_\sigma = 5$ is used (see Table~\ref{tab:ALS_nsigma}). All sampling and evaluation runs included a final denoising step based on Tweedie’s formula~\cite{robbins1992empirical}.

\paragraph{SDE Samplers.} 
For the SDE-based experiments, we employed the score networks from~\cite{song2021scorebasedsde}, namely NCSN++ (VE-SDE). The key hyperparameters in the unified SGM framework are the initial step size $\alpha$ and the sub-iteration parameter $n_\sigma$. Following the same strategy as in the NSCN (VE) experiments, we first performed partial FID evaluations to select the best $\alpha$, which was then fixed for the full FID runs. Similarly, $n_\sigma$ was tuned in the same manner; we observed that $n_\sigma = 2$ often yields the best performance in low-NFE settings.

For MC and LC correctors, we fixed $n_\sigma = 5$ throughout all experiments, as corrector-only samplers were already evaluated in the NSCN (VE) setting. Notably, our momentum corrector requires a smaller $\alpha$ than the Langevin corrector due to the use of our SNR schedule, where the step size is multiplied by $(1+\beta)^2$ during momentum-based sampling and evaluation.  

\paragraph{Observations.} 
We observe that the Langevin corrector improves predictor performance only in high-NFE regimes. In contrast, when the number of NFEs is small (e.g., 150 or 270), adding more Langevin correction steps worsens the FID. Finally, for predictor-only samplers such as EM or RD, these hyperparameters ($\alpha$ and $n_\sigma$) have no effect.

\begin{table*}[!ht]
    \label{tab:hyperparameter_NCSN++_cont}
    
    \centering
    \setlength{\tabcolsep}{2pt}
    \renewcommand{\arraystretch}{1.3}
    \caption{Used hyperparameters for NCSN++ (VE-SDE)  sampling/evaluation. We observe that the optimal $\alpha$ hyperparameter is, to some extent, linearly negatively correlated with the NFE.}
    \begin{tabular}{lccccc}
    \toprule
    \midrule
    Sampler & NFE $=150$  & NFE $=195$  & NFE $=210$ & NFE $=270$ & NFE $=300$\\
        \hline
    RD-MC & $\alpha=0.20,n_\sigma=2$& $\alpha=0.19,n_\sigma=2$& $\alpha=0.21,n_\sigma=2$ &  $\alpha=0.29,n_\sigma=2$& $\alpha=0.265,n_\sigma=2$\\
    RD-CDL & $\alpha=0.07,n_\sigma=2$& $\alpha=0.06,n_\sigma=2$& $\alpha=0.05,n_\sigma=2$ &  $\alpha=0.04,n_\sigma=2$& $\alpha=0.03,n_\sigma=2$ \\
    \midrule
         & NFE $=150$ & NFE $=270$ & NFE $=500$ & NFE $=1000$ & NFE $=2000$ \\
        \hline
        MC &$\alpha=0.30,n_\sigma=5$ & $\alpha=0.20,n_\sigma=5$& $\alpha=0.085,n_\sigma=5$ & $\alpha=0.05,n_\sigma=5$& $\alpha=0.0315,n_\sigma=5$\\
        LC &$\alpha=0.44,n_\sigma=5$ & $\alpha=0.38,n_\sigma=5$& $\alpha=0.32,n_\sigma=5$ &$\alpha=0.26,n_\sigma=5$&$\alpha=0.22,n_\sigma=5$\\
        CDL &$\alpha=0.12,n_\sigma=5$ & $\alpha=0.08,n_\sigma=5$& $\alpha=0.056,n_\sigma=5$ & $\alpha=0.039,n_\sigma=5$& - \\
        RD-LC & $\alpha=0.18,n_\sigma=1$& $\alpha=0.20,n_\sigma=1$& $\alpha=0.18,n_\sigma=1$ & $\alpha=0.18,n_\sigma=1$ &$\alpha=0.16,n_\sigma=1$\\
        
        EM-LC & $\alpha=0.24,n_\sigma=1$& $\alpha=0.18,n_\sigma=1$& $\alpha=0.18,n_\sigma=1$ & $\alpha=0.18,n_\sigma=1$ &$\alpha=0.16,n_\sigma=1$\\
    \bottomrule
    \end{tabular}
\end{table*}

\begin{table}[!ht]
    \centering
    \renewcommand{\arraystretch}{1.3}
    \setlength{\tabcolsep}{25pt}
    \caption{Hyperparameters for FFHQ, ImageNet (EDM, VE) sampling/evaluation.}
    \label{tab:my_label}
    \begin{tabular}{c|ccc}
    \toprule
    \midrule
        NFE & 150 & 300 & 450 \\
        \midrule
       FFHQ  &  $\alpha=0.08,n_\sigma=2$ & $\alpha=0.04,n_\sigma=2$ & $\alpha=0.027,n_\sigma=2$\\
       ImageNet & $\alpha=0.08,n_\sigma=2$ & $\alpha=0.04,n_\sigma=2$ & $\alpha=0.027,n_\sigma=2$ \\      
       Bedroom & $\alpha=0.08,n_\sigma=2$ & $\alpha=0.04,n_\sigma=2$ & $\alpha=0.027,n_\sigma=2$ \\  
       \bottomrule
    \end{tabular}
\end{table}

\section{Uncurated Samples}\label{app:samples}
\begin{figure*}[!h]
    \centering
    \includegraphics[width=0.8\linewidth,height=0.8\textwidth]{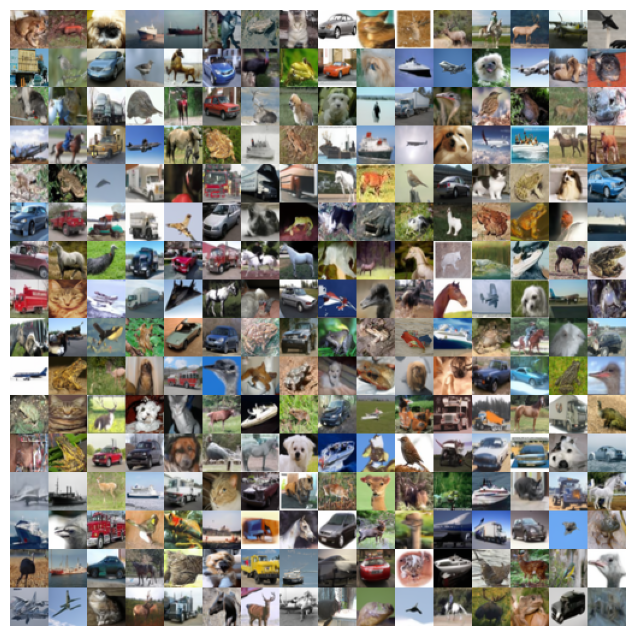}
    \caption{Additional samples on CIFAR10 from NCSN++ (VE-SDE)  using RD-MC sampler with NFE $=210$, we report an FID of $2.86$ with this setting.}
\end{figure*}

\begin{figure*}[!ht]
    \centering
    \includegraphics[width=0.8\linewidth,height=0.8\textwidth]{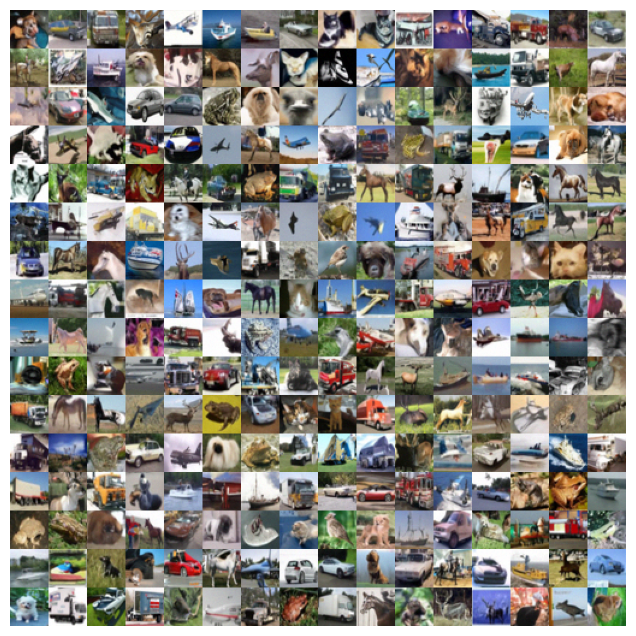}
    \caption{Additional samples on CIFAR10 from NCSN++ (VE-SDE)  using MC sampler with NFE $=500$, we report an FID of $9.16$ with this setting.}
\end{figure*}

\begin{figure*}[!ht]
    \centering
    \includegraphics[width=0.8\linewidth,height=0.8\textwidth]{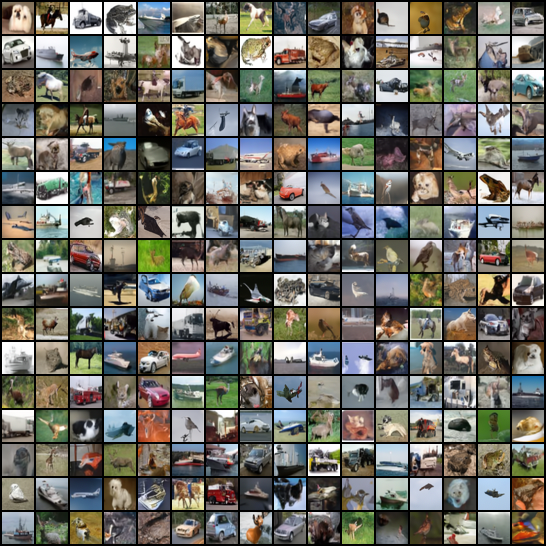}
    \caption{Additional samples on CIFAR10 from NSCN  using AMS sampler with NFE $=1000$. We report an FID of $9.82$ with this setting.}
\end{figure*}

\begin{figure*}[!ht]
    \centering
    \includegraphics[width=0.8\linewidth,height=0.8\textwidth]{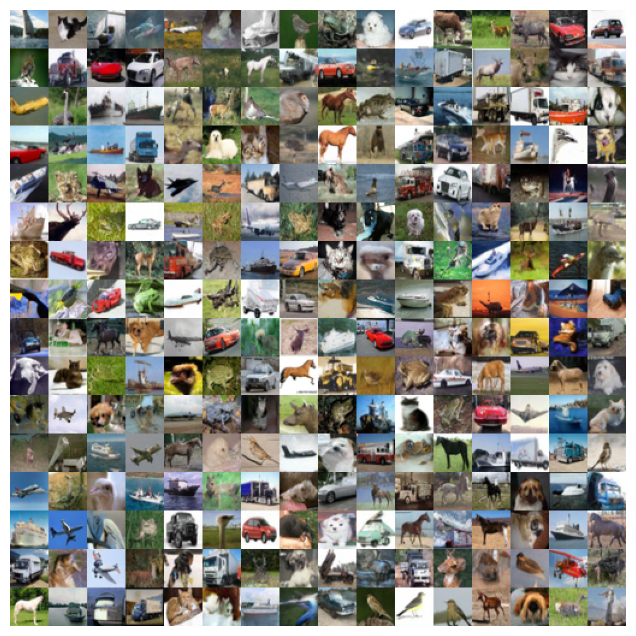}
    \caption{Additional samples on CIFAR10 from DDPM++ cont. (VP-SDE) using RD-MC sampler with NFE $=2000$.}
\end{figure*}

\begin{figure*}[!ht]
    \centering
    \includegraphics[width=0.8\linewidth,height=0.8\textwidth]{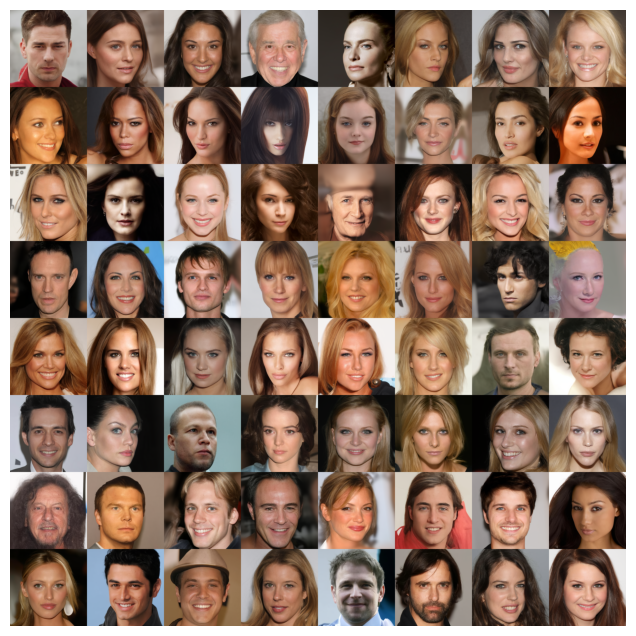}
    \caption{Additional samples on $256 \times 256$ CelebA-HQ from NCSN++ (VE-SDE)  using RD-MC sampler with NFE $=450$.}
\end{figure*}

\begin{figure*}[!ht]
    \centering
    \includegraphics[width=0.8\linewidth,height=0.8\textwidth]{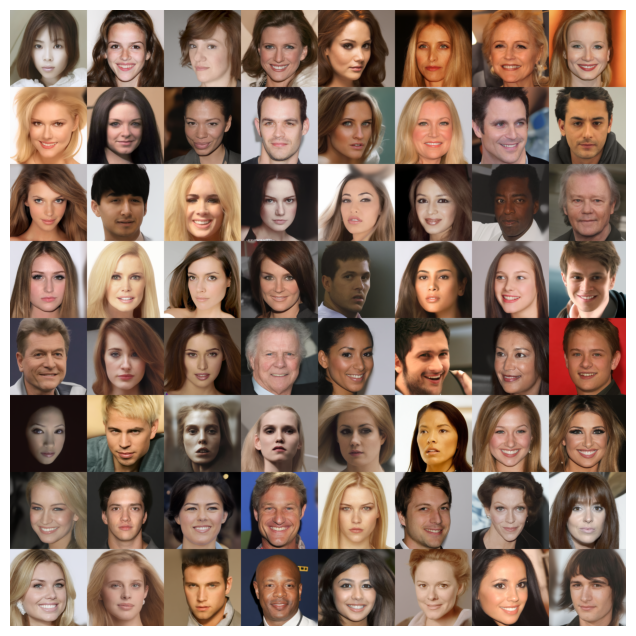}
    \caption{Additional samples on $256 \times 256$ CelebA-HQ from NCSN++ (VE-SDE)  using RD-MC sampler with NFE $=900$.}
\end{figure*}

\begin{figure*}[!ht]
    \centering
    \includegraphics[width=0.8\linewidth,height=0.8\textwidth]{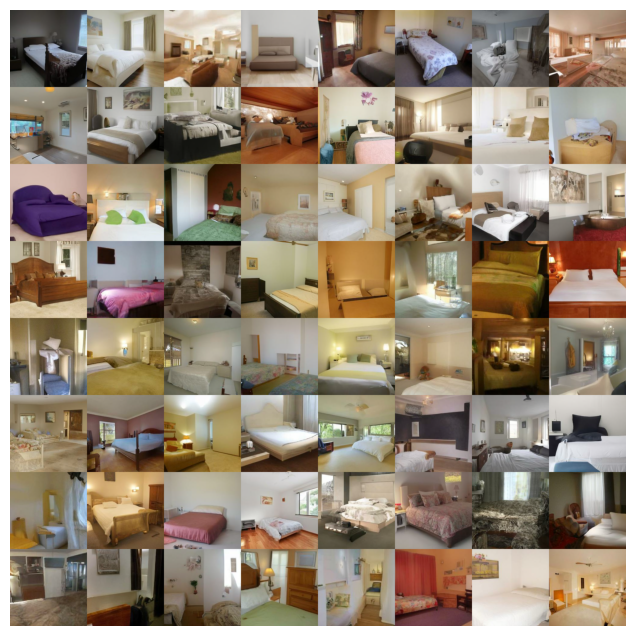}
    \caption{Additional samples on $256 \times 256$ bedroom from NCSN++ (VE-SDE)  using RD-MC sampler with NFE $=500$.}
\end{figure*}

\begin{figure*}[!ht]
    \centering
    \includegraphics[width=0.8\linewidth,height=0.8\textwidth]{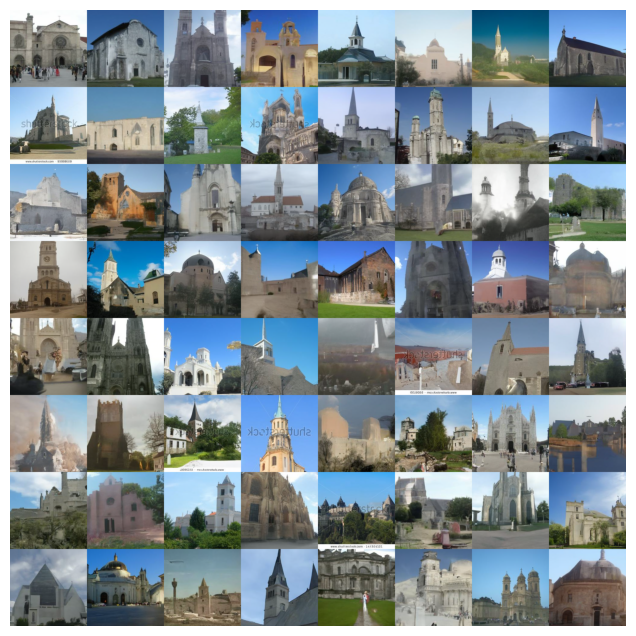}
    \caption{Additional samples on $256 \times 256$ church from NCSN++ (VE-SDE)  using RD-MC sampler with NFE $=300$.}
\end{figure*}

\begin{figure*}[!ht]
    \centering
    \includegraphics[width=0.8\linewidth,height=0.8\textwidth]{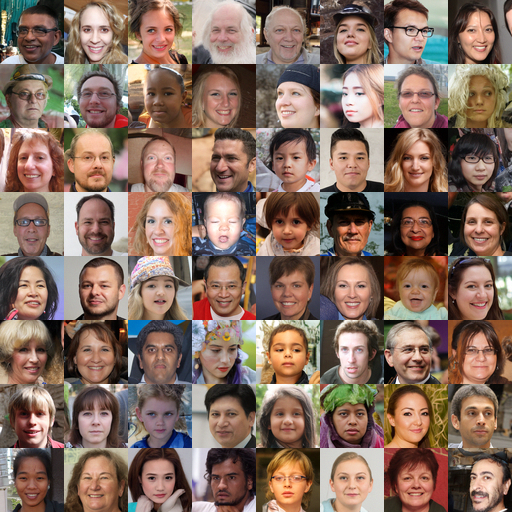}
    \caption{Additional samples on $64 \times 64$ FFHQ from EDM with VE schedule using RD-MC sampler with NFE $=450$.}
\end{figure*}

\begin{figure*}[!ht]
    \centering
    \includegraphics[width=0.8\linewidth,height=0.8\textwidth]{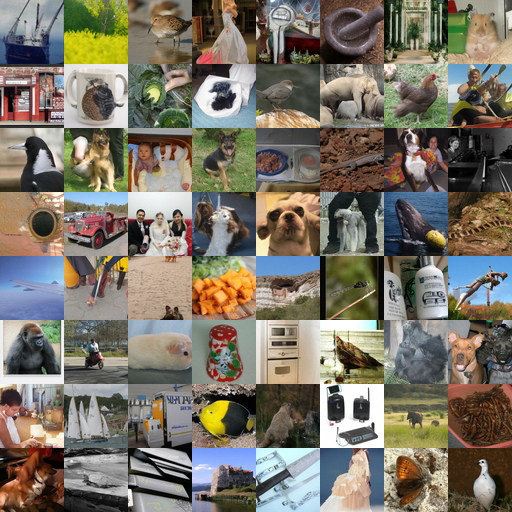}
    \caption{Additional samples on $64 \times 64$ ImageNet from EDM with VE schedule using RD-MC sampler with NFE $=450$.}
\end{figure*}